\documentclass[letterpaper, 10 pt, conference]{ieeeconf}  

\IEEEoverridecommandlockouts                              

\usepackage{graphics} 
\usepackage{kotex}
\usepackage{graphicx}
\usepackage{subcaption}
\usepackage{cite}
\usepackage{romannum}
\usepackage{amsmath,amssymb}
\usepackage{url}

\newcommand{\vs}{{\bf s}}
\newcommand{\va}{{\bf a}}
\newcommand{\vo}{{\bf o}}
\newcommand{\vx}{{\bf x}}

\usepackage{algorithm}
\usepackage[noend]{algpseudocode}
\makeatletter
\let\OldStatex\Statex
\renewcommand{\Statex}[1][3]{%
  \setlength\@tempdima{\algorithmicindent}%
  \OldStatex\hskip\dimexpr#1\@tempdima\relax}
\renewcommand{\ALG@beginalgorithmic}{\normalsize}
\makeatother

\DeclareCaptionFont{mysize}{\fontsize{8}{9.6}\selectfont}
\begin{document}
\title{\LARGE \bf
An Open-Source Low-Cost Mobile Robot System with an RGB-D Camera and Efficient Real-Time Navigation Algorithm}

\author{Taekyung Kim$^{1,*}$, Seunghyun Lim$^{2,*}$, Gwanjun Shin$^{3}$, Geonhee Sim$^{3}$, and Dongwon Yun$^{2,\dagger}$, \textit{Senior Member, IEEE}
\thanks{$^{*}$These authors contributed equally}
\thanks{$^{1}$T. Kim is with the Ground Technology Research Institute, Agency for Defense Development, Daejeon, 34186 Republic of Korea. Email: \texttt{ktk1501@add.re.kr}}%
\thanks{$^{2}$S. Lim and D. Yun are with the Department of Robotics Engineering, DGIST, Daegu, 30118 Republic of Korea. Email: \texttt{\{shl0216, mech\}@dgist.ac.kr}}%
\thanks{$^{3}$G. Shin and G. Sim are with the College of Transdisciplinary Studies, DGIST, Daegu, 42988 Republic of Korea. Email: \texttt{\{shinkansan, kh6362\}@dgist.ac.kr}}%
\thanks{$^{\dagger}$Corresponding author}%
}

\maketitle
\thispagestyle{empty}
\pagestyle{empty}

\begin{abstract}
Currently, mobile robots are developing rapidly and are finding numerous applications in the industry. However, several problems remain related to their practical use, such as the need for expensive hardware and high power consumption levels. In this study, we build a low-cost indoor mobile robot platform that does not include a LiDAR or a GPU. Then, we design an autonomous navigation architecture that guarantees real-time performance on our platform with an RGB-D camera and a low-end off-the-shelf single board computer. The overall system includes SLAM, global path planning, ground segmentation, and motion planning. The proposed ground segmentation approach extracts a traversability map from raw depth images for the safe driving of low-body mobile robots. We apply both rule-based and learning-based navigation policies using the traversability map. Running sensor data processing and other autonomous driving components simultaneously, our navigation policies perform rapidly at a refresh rate of 18Hz for control command, whereas other systems have slower refresh rates. Our methods show better performances than current state-of-the-art navigation approaches within limited computation resources as shown in 3D simulation tests. In addition, we demonstrate the applicability of our mobile robot system through successful autonomous driving in an indoor environment.
\end{abstract}

\section{Introduction}
\label{sec:introduction}
Recently, mobile robots have been navigating cluttered environments such as buildings and roads. Implementation of these devices into various industrial fields has been accelerating \cite{liu2020robot}. Depending on the design and purpose, they are utilized in various areas, such as for delivery, guidance, searches, and inspections. Therefore, robot navigation in crowded environments has been studied as a key topic in many research fields. Furthermore, the demand for mobile robots is increasing not only in industrial fields but also for individual uses. Examples include social robots, home service robots, and cleaning robots.

However, expensive hardware and high power consumption are hindering the practical application of mobile robots \cite{mei2004energy}. For safe driving, the ability to recognize traversal areas and to detect obstacles is critical in an advanced motion planning strategy. LiDARs have been used as a dominant sensor to ensure accurate distance measurements and have been combined with cameras for deep learning recognition. However LiDARs are far more expensive than other sensors and thus increase the price of the robot. Simultaneous localization and mapping (SLAM) \cite{durrant2006simultaneous}, mainly used for indoor positioning, requires a high-performance computer. Meanwhile, with the rapid development of deep reinforcement learning (DRL), numerous studies have focused on the use of neural networks for robot navigation \cite{chen2017decentralized, everett2018motion, chen2019crowd}. Although the inference of a DRL model can be done in a short time, environment recognition requires heavy iterations for sensor data processing. This should be supported by a high performance computer, and the process ends up draining the battery more rapidly.

\begin{figure}[t]
\centering
\subfloat[]{
\includegraphics[width=0.8\linewidth]{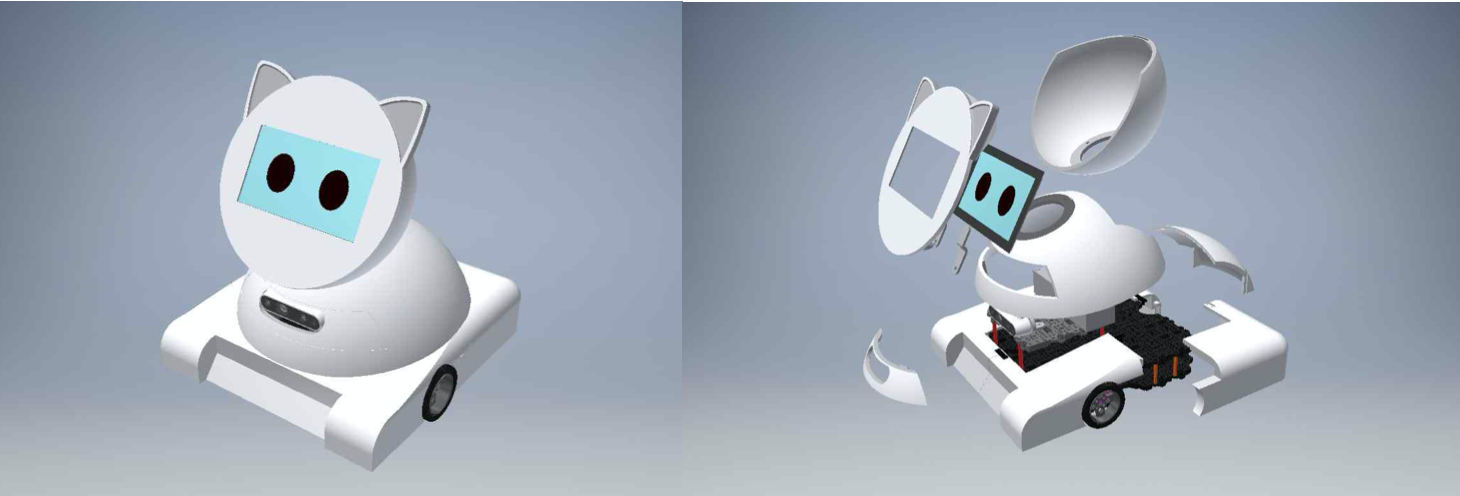}
\label{subfig:3DModel}
}
\qquad
\subfloat[]{
\includegraphics[width=0.8\linewidth]{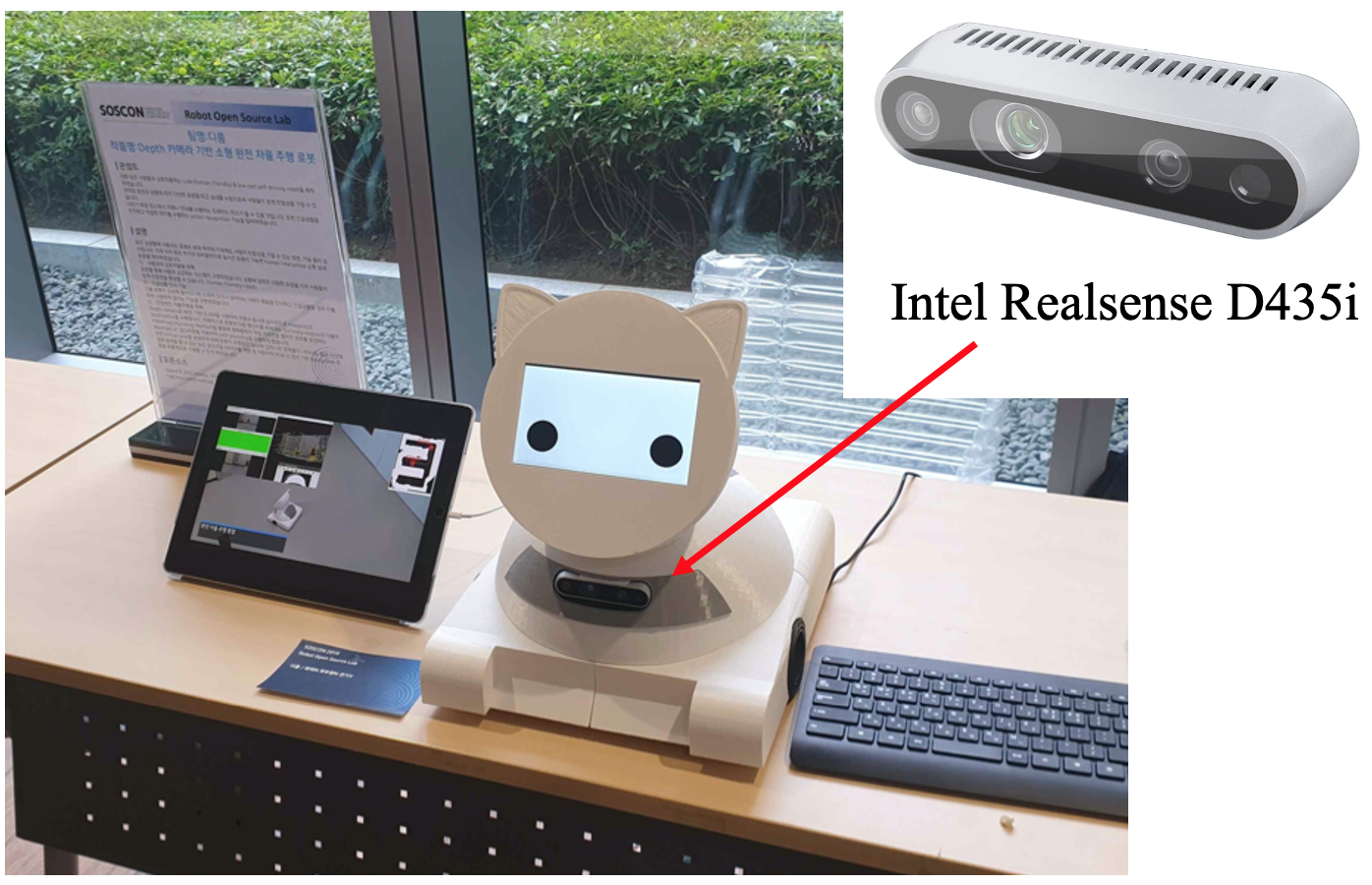}
\label{fig:RealDPoom}
}
\caption{Our indoor mobile robot platform: DPoom. (a) 3D model of the robot chassis and components. (b) Image showing the actual appearance of DPoom.}
\label{fig:DPoom}
\end{figure}

Furthermore, software components for environmental interactions such as image-based object detection should be followed by navigation components to serve a socially interactive robot. These image-based recognition algorithms rely highly on the graphics processing unit (GPU). Therefore, navigation algorithms with low GPU utilization are greatly welcomed in mobile robot systems, even if they have an on-board GPU.

We build an open-source low-cost autonomous mobile robot system without a need for a high-performance GPU or LiDARs that successfully overcomes the aforementioned problems. We also propose a real-time navigation approach designed for low-cost indoor mobile robots. Only an RGB-D camera is used for environment recognition, and real-time performance is achieved on a low-end single-board computer (SBC) without external computing aids. The robot can build a point cloud map and perform real-time positioning by means of lightweight RGB-D SLAM. We use a modified A* algorithm that generates a stable path while maintaining sufficient distances from adjacent obstacles. In addition, we propose a ground segmentation approach that provides a compact traversability map in real time using an RGB-D camera. This approach enables the robot to navigate among pedestrians safely. We demonstrate the feasibility of our ground segmentation method using both  rule-based and learning-based navigation policies with the traversability information. All of the software for fully autonomous driving is integrated on our mobile robot platform, DPoom (see Fig.~\ref{fig:DPoom}). For human-computer interactions, friendly expressions are displayed on the front screen. It also has an appropriate exterior design for educational and socially interactive purposes. We deployed DPoom as a social robot in a crowded residential environment. All of our materials, including the hardware and software, are released under an open-source license.\footnote{
Our entire works including hardware and software are in: \url{https://github.com/shinkansan/2019-UGRP-DPoom}. 
Our detailed video is available in: \url{https://youtu.be/Li3-RlO28lk}.
}



\section{Related Work}
\label{sec:related work}
System design of a robot is important, as the design determines its purpose, function and price. Most traveling mobile robots are designed for mission automation. For full automation, several essential functions should be realized synchronously with cross-interaction capabilities. In modern autonomous driving, the task is developed with separately divided modules that are integrated in a pipeline. Localization is the most basic module for all control tasks with closed-loop feedback. SLAM is generally used for indoor robot localization. In order to drive to a certain location in a wide area, the robot should generate a trajectory through global path planning via, for instance, the A* algorithm \cite{Hart1968}. When obstacles not on the prior map or moving objects appear on the planned trajectory, the robot avoids them by motion planning.

Collision avoidance and safe navigation are particularly important for stable robot operation \cite{haddadin2009requirements}. Reciprocal velocity obstacles \cite{van2008reciprocal} and optimal reciprocal collision avoidance (ORCA) \cite{van2011reciprocal} have been commonly used for dynamic robot navigation. However, given that they are based on hand-crafted functions, they do not work well in more complex environments. Recent works applied DRL to navigation in crowds \cite{chen2017decentralized}, \cite{everett2018motion}, \cite{chen2017socially}. These approaches assume that the robot is aware of objects in a $360\,^{\circ}$ field of view (FOV) and that it accurately measures the positions of objects in real time with LiDAR. Unlike those assumptions, point cloud processing is computationally expensive and lowers the decision frequency of navigation algorithm when running on an onboard computer. A slow decision often causes frozen robot situations or even collisions. Therefore, it is necessary to choose a navigation policy that guarantees real-time execution according to the robot's computational performance. Furthermore, if procuring a $360\,^{\circ}$ FOV considering the assumptions above, the price of the required sensor increases greatly. This also places negative constraints on the mechanical design and on the design of the robot body's exterior components. Meanwhile, successful navigation is coupled with the ability to estimate traversable areas, not merely depending on the navigation policy. Recent ground segmentation methods based on a convolutional neural network (CNN) incur a high computational cost \cite{yang2019robustifying}, \cite{paigwar2020gndnet}.

In this paper, we use an RGB-D camera, which is generally much less expensive than a 3D LiDAR. The depth data provide robustness for localization and the direct distances to obstacles without estimating them with heavy algorithms. The robot body design was convenient because the sensor does not need to be mounted on top and an empty layer is not required in the middle of the body for laser range scanning. Real-time navigation is possible using our RGB-D ground segmentation approach.

\section{Approach}
\label{sec:approach}
\subsection{DPoom indoor mobile robot platform}
DPoom (see Fig.~\ref{fig:DPoom}) is an open-source indoor autonomous mobile robot designed to interact with people while traveling around indoor environments. It was developed while focusing on three factors: cost performance, human-robot interaction, and ease of use.

DPoom is built for fully autonomous driving using only a low-end SBC (LattePanda Alpha 864, LattePanda) and an RGB-D camera (Realsense D435i, Intel). The low-end SBC consists of an Intel dual-core m3-8100y processor, 8 \textit{GB} RAM and Intel HD 615 on-board graphics. The controller board (OpenCR, ROBOTIS) for our system is is embedded with a robot operating system (ROS) \cite{quigley2009ros} and has a nine-axis IMU Sensor MPU9250. The robot uses the front RGB-D camera D435i with a 1280$\times$720 resolution to recognize the environment. The camera has $85.2\,^{\circ}$ $\times$ $58\,^{\circ}$ FOV. Two actuators (Dynamixel XM430-W210-T, ROBOTIS) are used for wheel driving, and two ball casters at the bottom of the rear structure support differential driving. The software interface is built on Ubuntu-based ROS Kinetic. The hardware is 33.0$\times$33.5$\times$35.0 \textit{mm} (width$\times$depth$\times$height), weighs approximately 4 \textit{kg}, and can achieve speeds up to 0.26 \textit{m/s}.

\subsection{3D SLAM}
Mapping should be preceded before deploying a robot to an unknown area. Mapping enables the robot to plan its trajectory to the goal and to perform localization to estimate its pose. 3D mapping using RGB-D data is known to be capable of higher accuracy than monocular vision-based mapping as it provides additional robust features for scan matching. We used RTAB-Map \cite{labbe2019rtab} to perform mapping and localization simultaneously with RGB-D data and wheel odometry. RGB image frames and depth frames were obtained from the Intel Realsense D435i with Intel Realsense SDK 2.0. Wheel-based odometry was calculated in the OpenCR controller using data from the built-in IMU sensor and motor encoders.

\begin{figure}[t] 
\centering
\subfloat[]{\includegraphics[width=0.23\textwidth]{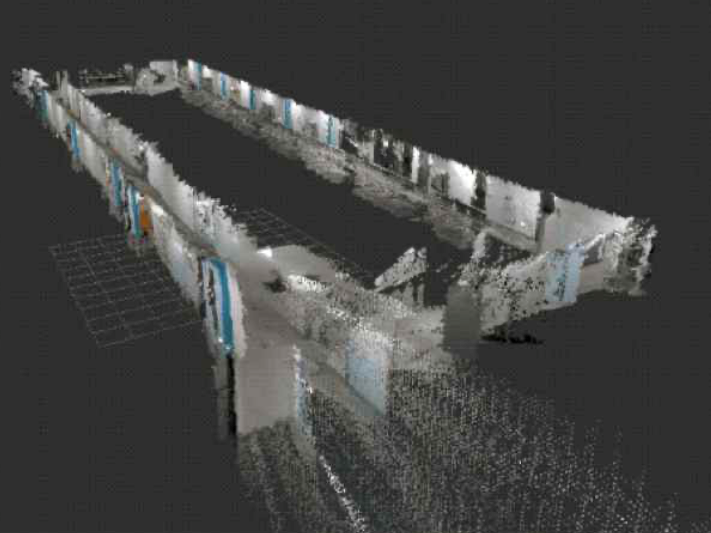}} \hspace*{0.01\textwidth}
\subfloat[]{\includegraphics[width=0.23\textwidth]{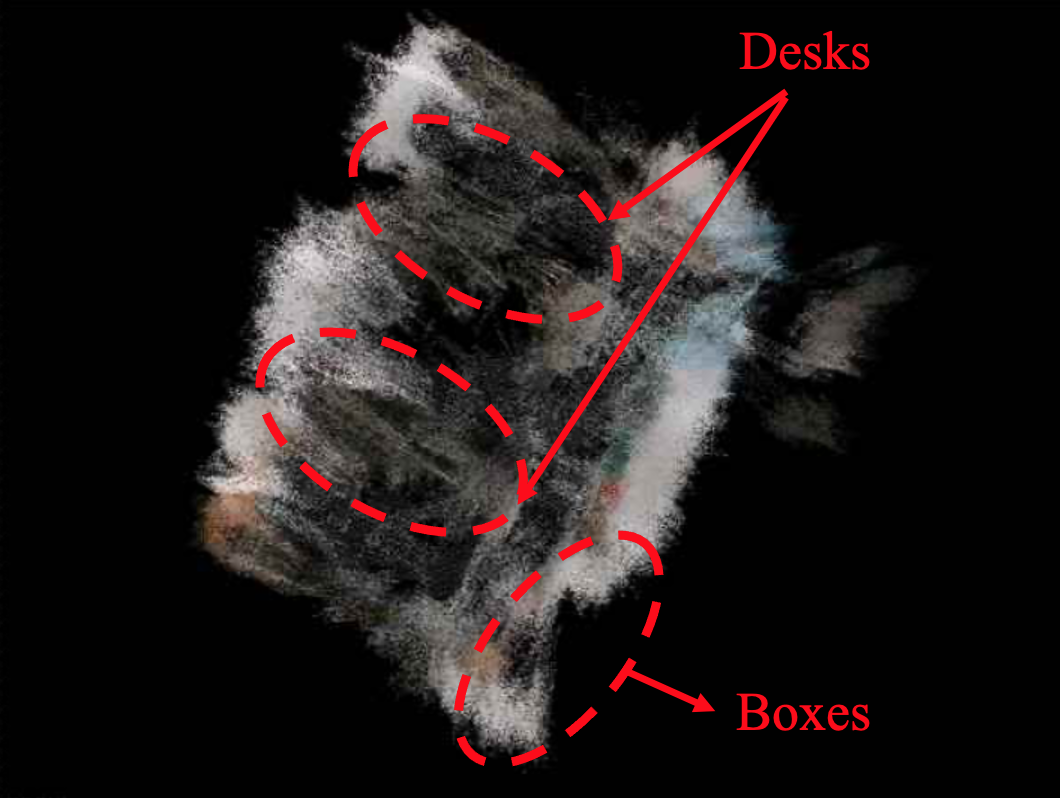}} \hfill
\subfloat[]{\includegraphics[width=0.23\textwidth]{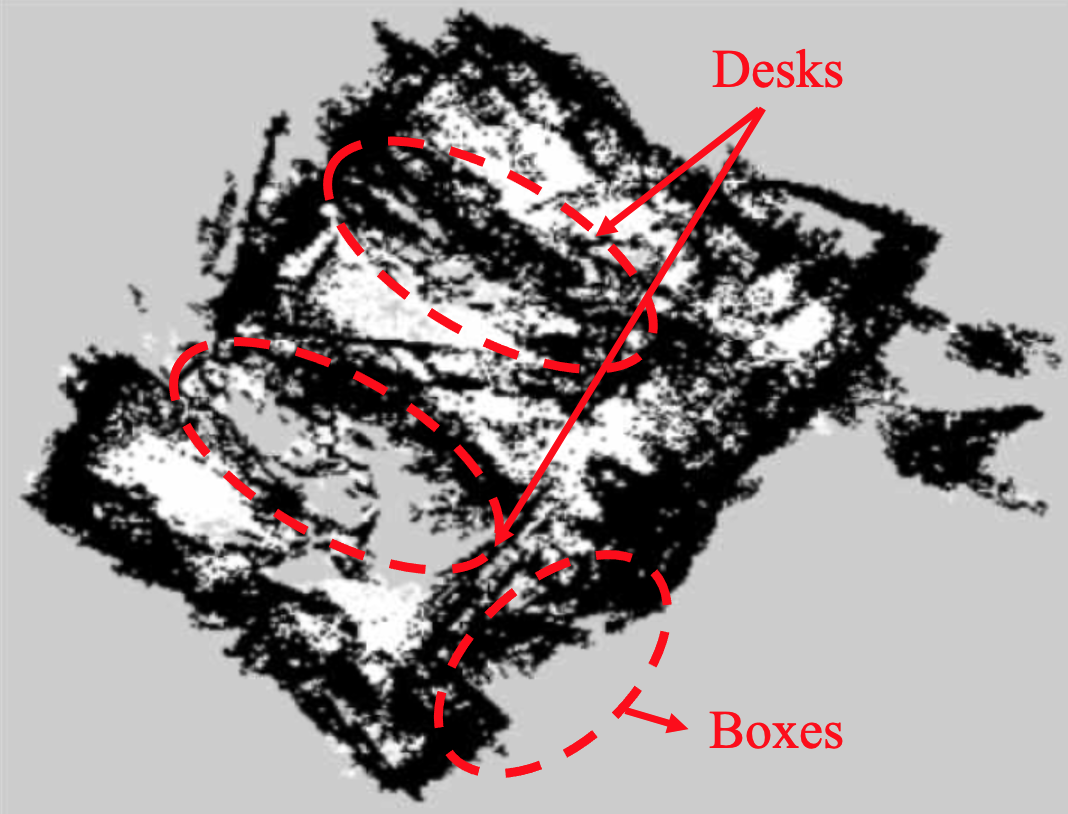}} \hspace*{0.01\textwidth}
\subfloat[]{\includegraphics[width=0.23\textwidth]{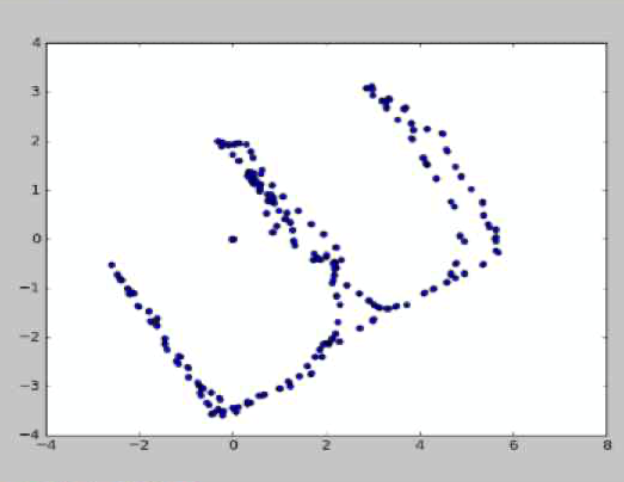}} \hfill
    \caption{Environment representation and localization of 3D SLAM. (a) Point cloud map after 3D mapping around the floor area. (b) Bird's-eye view of a point cloud map for an indoor room. (c) 2D occupancy grid map through the processing of 3D point cloud projection. (d) Robot trajectory in the localization mode.}
    \label{fig:SLAM}
\end{figure}

Environments are represented as a point cloud map or a grayscale occupancy grid map after SLAM. Fig.~\ref{fig:SLAM}a presents the result when mapping around a building hallway. Point cloud maps are saved in local memory and are used for scan matching during localization. The projected occupancy grid map can be used as prior knowledge of global path planning. The localization results are synchronously published to the ROS middleware in our system.

\subsection{Global path planning}
Under a 2D environment, the deterministic planning can generate more accurate paths in less time than a probabilistic path planner \cite{Khanmirza2018}. In this study, we use the A* algorithm \cite{Hart1968}, a deterministic path planning algorithm, as the global path planner. The A* algorithm searches for the path by adding information about the goal node to the Dijkstra algorithm \cite{dijkstra1959}, which is the most basic path planner. 



By introducing the distance cost \(d(n)\) to the cost calculation of the A* algorithm as shown in Equation~\ref{eq:modified}, it is possible to generate a path with more stability, maintaining a proper distance from obstacles \cite{Lim2020}. We used the fast marching method (FMM) to calculate the distance cost, as this approach solves the boundary value problem of the Eikonal equation \cite{Sethian2016}. The distance cost is designed to have a larger value as the node becomes closer to nearby obstacles. Hence, the modified A* algorithm generates a smoothed path with a tendency to keep a distance from nearby obstacles.
\begin{equation} \label{eq:modified}
    f(n) = g(n) + h(n) + d(n).
\end{equation}

\subsection{Fast raw depth image ground segmentation}
When a low floor mobile robot follows a globally planned path, it is necessary not only to detect obstacles for collision avoidance but also to recognize whether the floor surface is traversable. Along with the rapid growth in the field of deep learning, research on ground traversability estimation with RGB-D cameras has been also conducted \cite{schilling2017geometric}. Yang et al. demonstrated the robustness of CNN-based models \cite{yang2019robustifying}. Paigwar et al. presented a real-time preprocessing method and a CNN model for application to robot navigation \cite{paigwar2020gndnet}. However, these methods require GPUs for real-time operation. In addition, deep-learning-based approaches can take into account floor thresholds, but they are unable to adjust the height of ground threshold on a deployed model depending on the situation. Floor thresholds are disastrous to low floor robots, as they can cause a malfunction or cause the robot to overturn. Therefore, there is a need for an analytic algorithm capable of adjusting the threshold height according to the robot platform and driving situation.

Mathematically derived estimation algorithms have been developed at the same time. For a mobile robot system designed at a low price point, the algorithm must be computationally efficient and must support real-time implementation. Holz et al. showed that real-time plane segmentation is possible on a CPU by clustering and merging points in normal space \cite{holz2011real}. However, this requires an additional analysis to determine whether each plane area is actually drivable when considering physical constraints such as the vehicle’s width and rotation angle.

\begin{figure}[t] 
\centering
\subfloat[]{\includegraphics[width=0.48\textwidth]{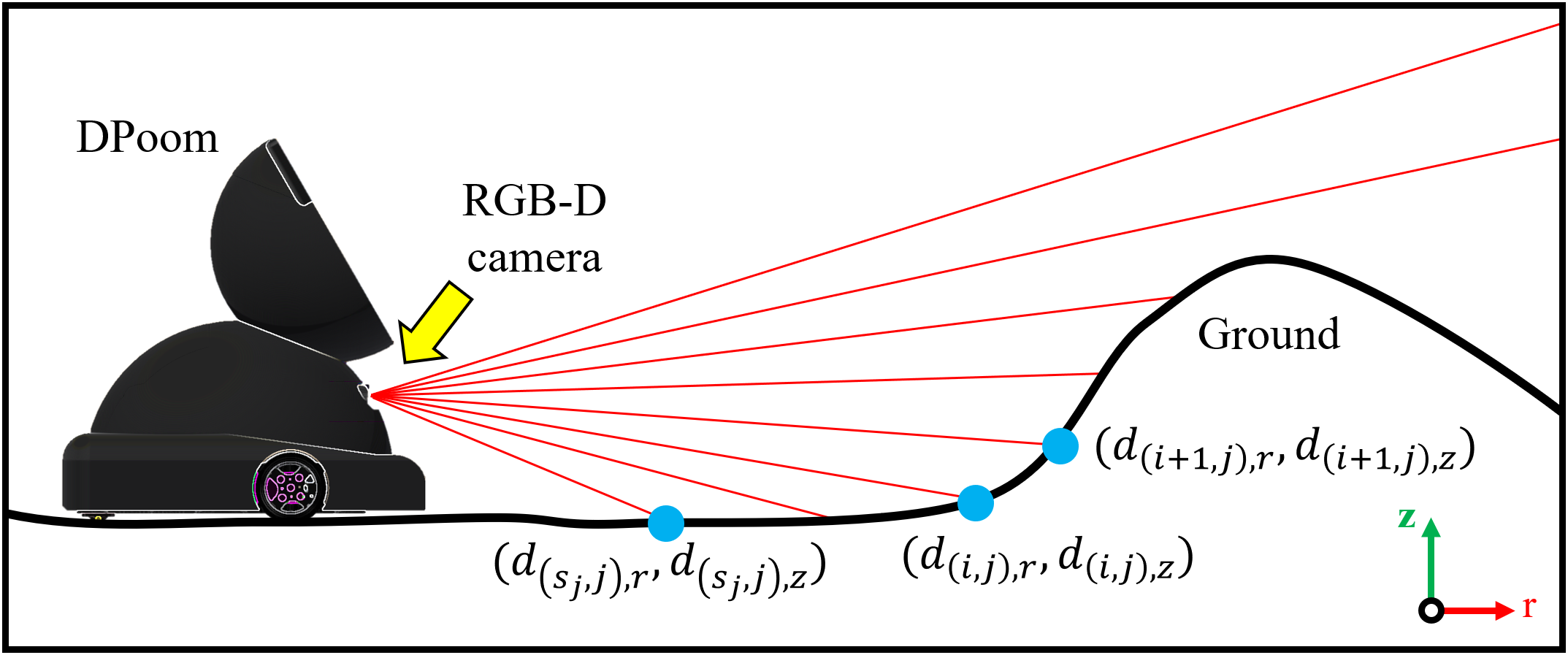}} \hfill
\subfloat[]{\includegraphics[width=0.23\textwidth]{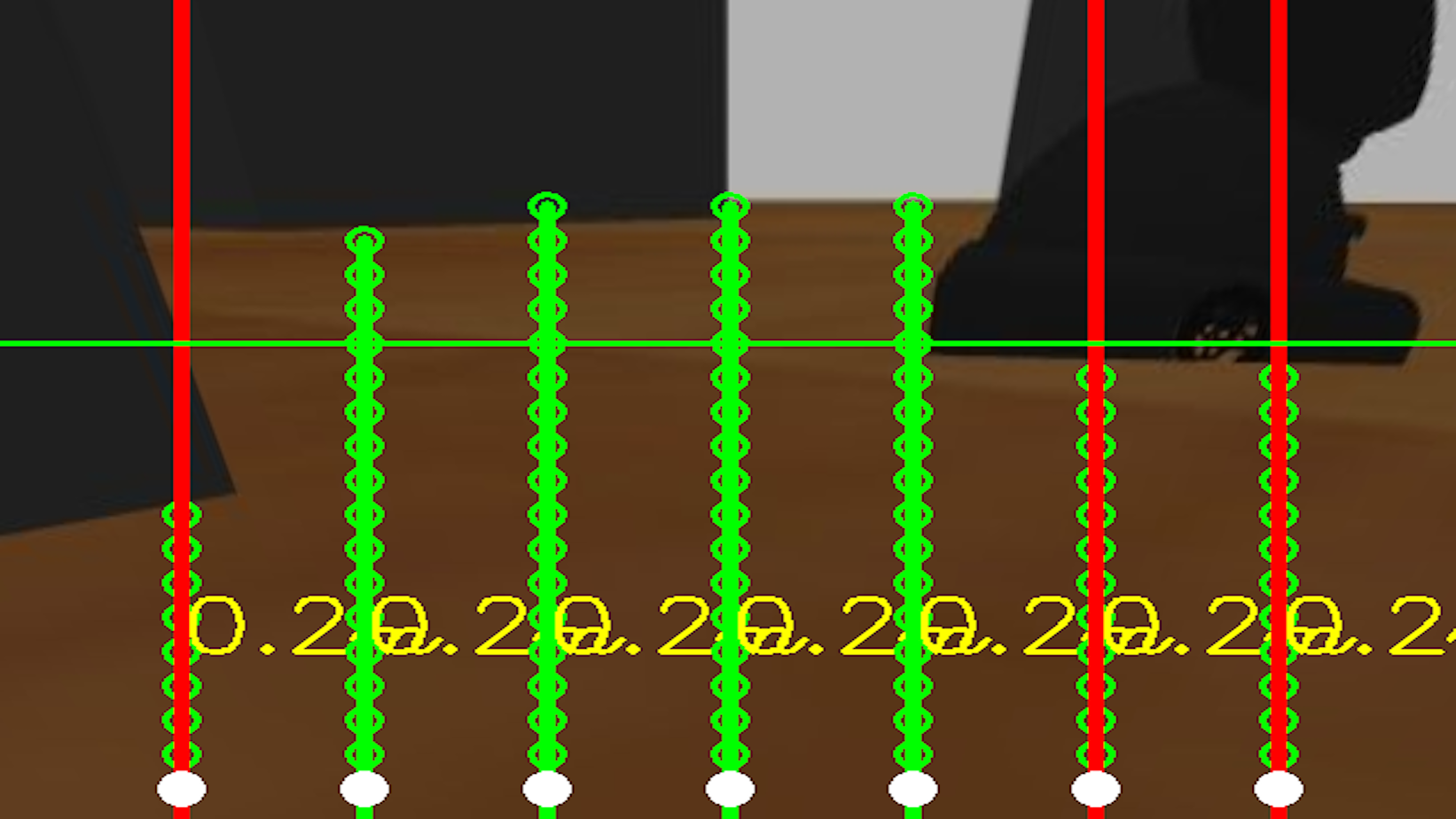}} \hspace*{0.01\textwidth}
\subfloat[]{\includegraphics[width=0.23\textwidth]{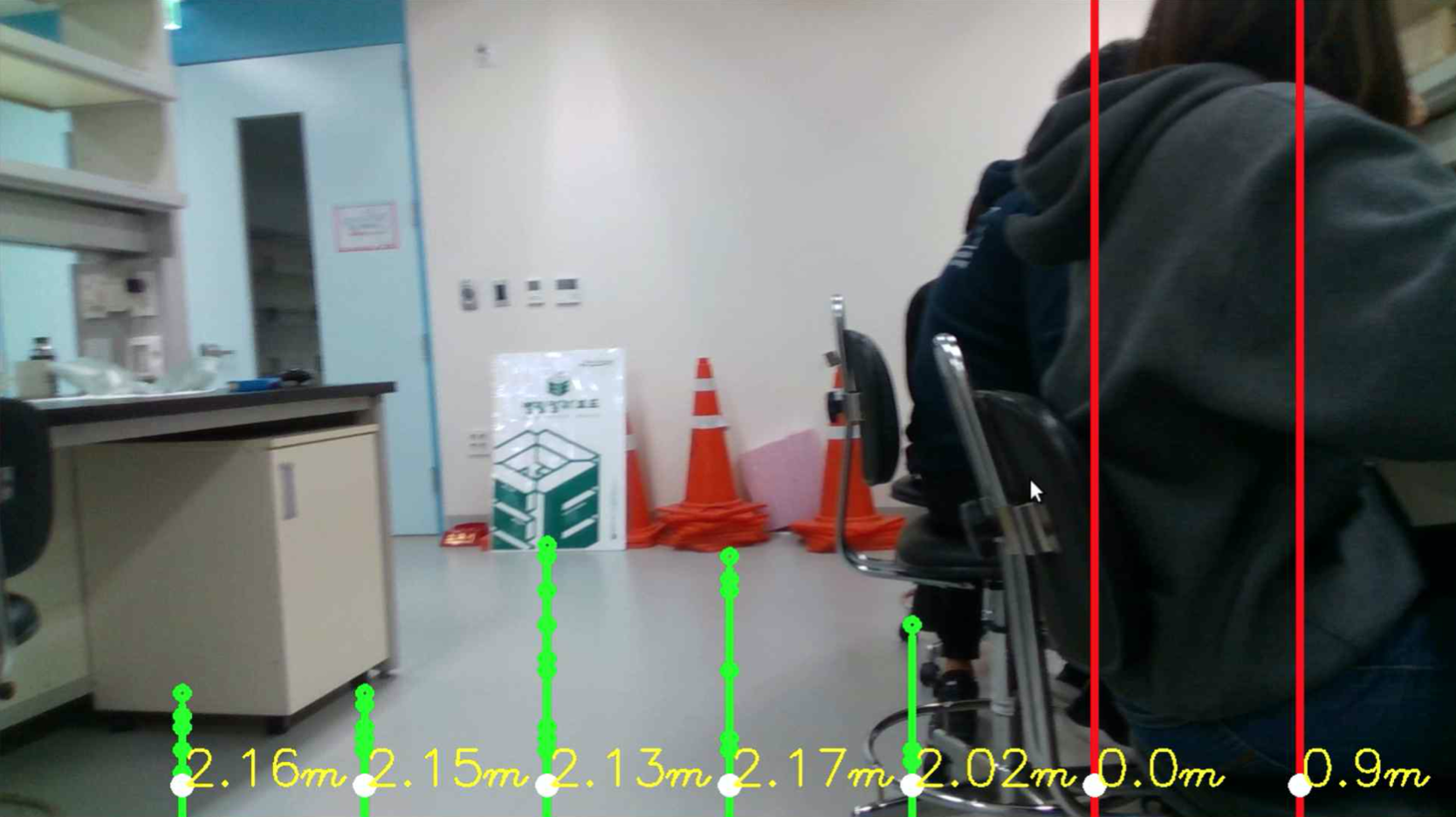}}  
\caption{Notations and results of the ground segmentation algorithm. (a) Notations for the ground segmentation algorithm in side view. (b) Our raw depth image ground segmentation result in the Gazebo environment. (c) Result in the real world.}
\label{fig:ground_seg}
\end{figure}

\begin{figure*}[t]
\centering
\vspace*{0.1in}
\includegraphics[width=0.95\linewidth]{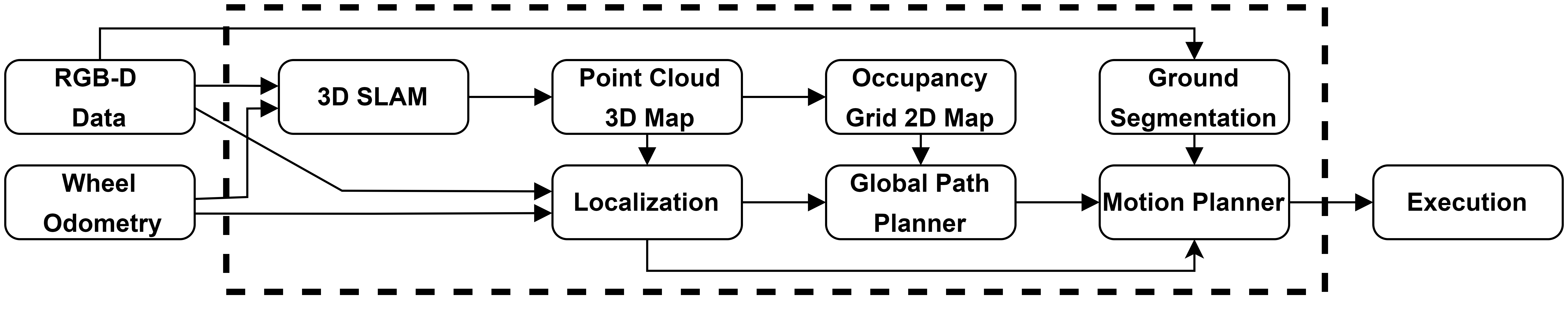}
\caption{Software architecture of our autonomous driving system.}
\label{fig:system}
\vspace*{-0.1in}
\end{figure*}

Here, we propose a concept known as \textit{milestone over rendered paths} (MORP), a real-time ground segmentation algorithm that can robustly recognize the forward traversal area with an RGB-D camera and that is designed to avoid obstacles effectively. With this algorithm, motion planning can be solved with a very small amount of computation by separating the area in front of the robot into virtual lanes and using the information of the closest non-traversable point recognized in each lane. Ground segmentation is performed for the center path of each lane area, and the first encountered non-traversable point is saved as a dead-end. This procedure is similar to 2D line extraction for fast segmentation of 3D point clouds \cite{himmelsbach2010fast}. A raw depth image is used for ground segmentation, which is converted from a depth point cloud into a 2D gray-scale image. Holz et al. showed that considering the pixel neighborhoods instead of spatial neighborhoods leads to a significant increase in the point cloud processing speed at the cost a small degree of accuracy \cite{holz2010towards}. It is possible to represent a large lane area by performing single column segmentation in a raw depth image. The score of each pixel is a weighted sum of the gradient from both the assigned start pixel depth and the adjacent pixel depth relative to the current pixel depth. Let $(w_1, w_2)$ denote the weighting factors of the gradients. The first pixel exceeding the score limit will be the dead-end in that area. Let ($d_{(i,j),r}, d_{(i,j), z}$) denote the actual position from the sensor of pixel $P_{(i,j)}$ in the N $\times$ M size depth image, as shown in Fig.~\ref{fig:ground_seg}. Let \(n\) denote the number of virtual lanes to scan. The dead-ends $\mathbb{D}(n)$ consist of the set of the largest indices to be segmented as the ground in each column \(j\):

\setlength{\belowdisplayskip}{0pt} \setlength{\belowdisplayshortskip}{0pt}
\setlength{\abovedisplayskip}{0pt} \setlength{\abovedisplayshortskip}{0pt}
\begin{equation}
\mathbb{D}(n) = \underset{j= \frac{M}{n} , \frac{2M}{n}, \dots } { \overset{1-\frac{M}{n}}{\bigcup}} \max \, ( \, \mathbf{S}_i(j) \, )
\end{equation}

\begin{align}
\mathbf{S}_i(j) & = \begin{aligned}[t]  \{ i \in (s_j, 
& s_{j+1} , \dots, N) \, | \,   w_1  \left( \frac{d_{(i+1,j),z} - d_{(i,j),z}}{d_{(i+1,j),r} - d_{(i,j),r}} \right) \\
& + w_2 \left( \frac{d_{(i+1,j),z} - d_{(s_j,j),z}}{d_{(i+1,j),r} - d_{(s_j,j),r}} \right) < C \}, 
\end{aligned}
\end{align}
where \(s_j\) is the start index of the ground on column \(j\), and \(C\) is the empirically determined threshold to segment as the ground. When this process is done sparsely over the entire image, milestones are generated for a traversability map containing robust but very compact data. It is easy to implement parallel computing because the process is explicit and the execution time is virtually consistent on each lane. The implementation results are visualized in Fig.~\ref{fig:ground_seg}.

More dense segmentation will be done on the front area of the robot as \(n\) becomes larger. However, there is a trade-off relationship between the density and the computation cost. An appropriate \(n\) should be selected in consideration of the width and the driving performance of the robot.

\subsection{Robot navigation based on traversable areas}
Given that each dead-end in $\mathbb{D}(n)$ contains direct information about non-traversable areas, collision avoidance is possible just with simple rule-based decisions. Avoidance in this case works in the same manner as a vehicle lane change on a road. On the other hand, this type of data pre-processing to obtain compact and representative features $\mathbb{D}(n)$ can be used as the observation space ($\vo$) for reinforcement learning. One powerful and sample-efficient approach of reinforcement learning is imitation learning. Behavior cloning (BC) in particular has been successfully used in many robotics applications given its simplicity and efficiency \cite{osa2018algorithmic}. In this work, we also implement a neural network policy based on BC to verify the feasibility of our lightweight ground segmentation method aside from rule-based navigation policy. Both policies are evaluated in Section~\ref{sec:experimental results}.

\begin{table}[t]
\vspace*{0.1in}
\renewcommand\arraystretch{1.15}
\captionsetup{justification=centering}
\caption{}
\caption*{\textsc{Overall architecture of the neural network policy}}
\label{table:model}
\setlength{\tabcolsep}{3pt}
\begin{center}
\begin{tabular}{c|cc}
\hline
\, \, &  Size & Activation   \\ \hline
\,Input\,      & $(\left|\vs\right| + \left|\vo\right| + \left|\va\right|) \times H$         &  -      \\
\,Hidden Layer 1\,     & $2 \times$Input Size        &  ReLU~\cite{nair2010rectified}      \\
\,Hidden Layer 2\,  & $3 \times$Input Size      &  ReLU     \\
\,Hidden Layer 3\,      & $1 \times$Input Size    &  Linear    \\
\,Output\,      & $\left|\va\right|$   &  -    \\
\hline
\end{tabular}
\end{center}
\vspace*{-0.1in}
\end{table}

We used a feedforward fully-connected neural network for imitation learning. Here, we denote the robot position relative to the goal point as $x$ and $y$, and denote the robot direction as $\theta$. The robot state is defined as $\vs = [x, y, \theta]$. The action command is defined as $\va = [v_x, \omega_z]$, representing the longitudinal velocity and angular velocity of the robot, respectively. Then, we can consider a policy $\hat{\pi}$ that takes $\vx_t = [\vs_t, \vo_t, \va_t]$ at time $t$ as input and results in the next action $\va_{t+1}$ as output. If the policy is given only a single time step information, capturing high-level intentions from the human demonstrations may be ambiguous. Therefore, we provide the state, observation and action history here as the input of the policy. In this case, the observation history contains implicit information about the velocities of the moving obstacles. This approach has been introduced in relation to applications of helicopters, autonomous vehicles, and quadruped legged robots \cite{punjani2015deep, spielberg2019neural, kumar2021rma}. The length of the history is denoted as $H$. The overall structure of our neural network policy $\va_{t+1} = \hat{\pi}(\vx_{t-H+1}, \vx_{t-H+2}, \dots \vx_{t})$ is shown in Table~\ref{table:model}.

Finally, all modules are synchronously integrated into the autonomous driving system. In order to navigate to the goal in a complex environment, the ideal trajectory should be generated from global path planning as part of the proposed method. The orientation toward the way point should be updated while avoiding obstacles via localization. A real-time traversability analysis is required to avoid local obstacles successfully. The overall system architecture is shown in Fig.~\ref{fig:system}.


\section{Experimental results}
\label{sec:experimental results}

\subsection{3D SLAM}
\begin{figure}[t] 
\centering
\subfloat[]{\includegraphics[width=0.22\textwidth]{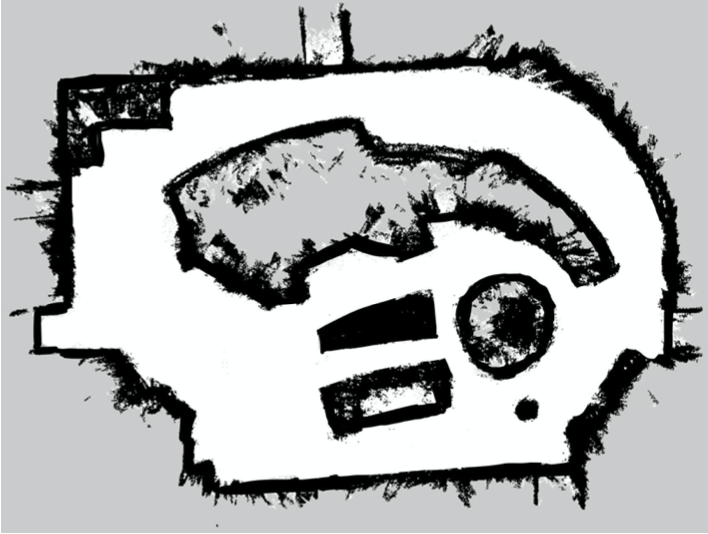}} \hspace*{0.01\textwidth}
\subfloat[]{\includegraphics[width=0.24\textwidth]{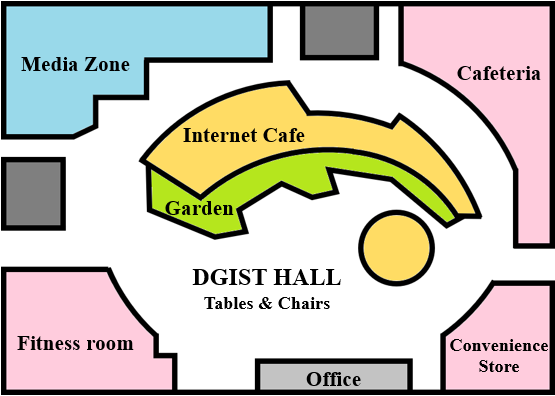}} 
    \caption{Result of 3D mapping and 2D projection. (a) 2D projection of the point cloud map. (b) Actual floor plane of the area.}
\label{fig:slam_result}
\end{figure}

We tested our SLAM in a residential lobby on the DPoom platform. The 3D point cloud map was saved in local memory for localization. Fig.~\ref{fig:slam_result}a depicts the obtained occupancy grid map. We added artificial grass and tables as confined areas on the map. Compared with Fig.~\ref{fig:slam_result}b, the map shows that a loop closure was performed to correct distortion and elevation issues. 

\subsection{Global path planning}
The modified A* algorithm uses a binarized 2D occupancy grid map. It is also used to obtain a distance cost map by FMM. The modified A* algorithm generates a path by calculating the fitness cost based on the binarized map and the distance cost map.
    
\begin{figure}[t] 
\centering
\subfloat[]{\includegraphics[width=0.23\textwidth]{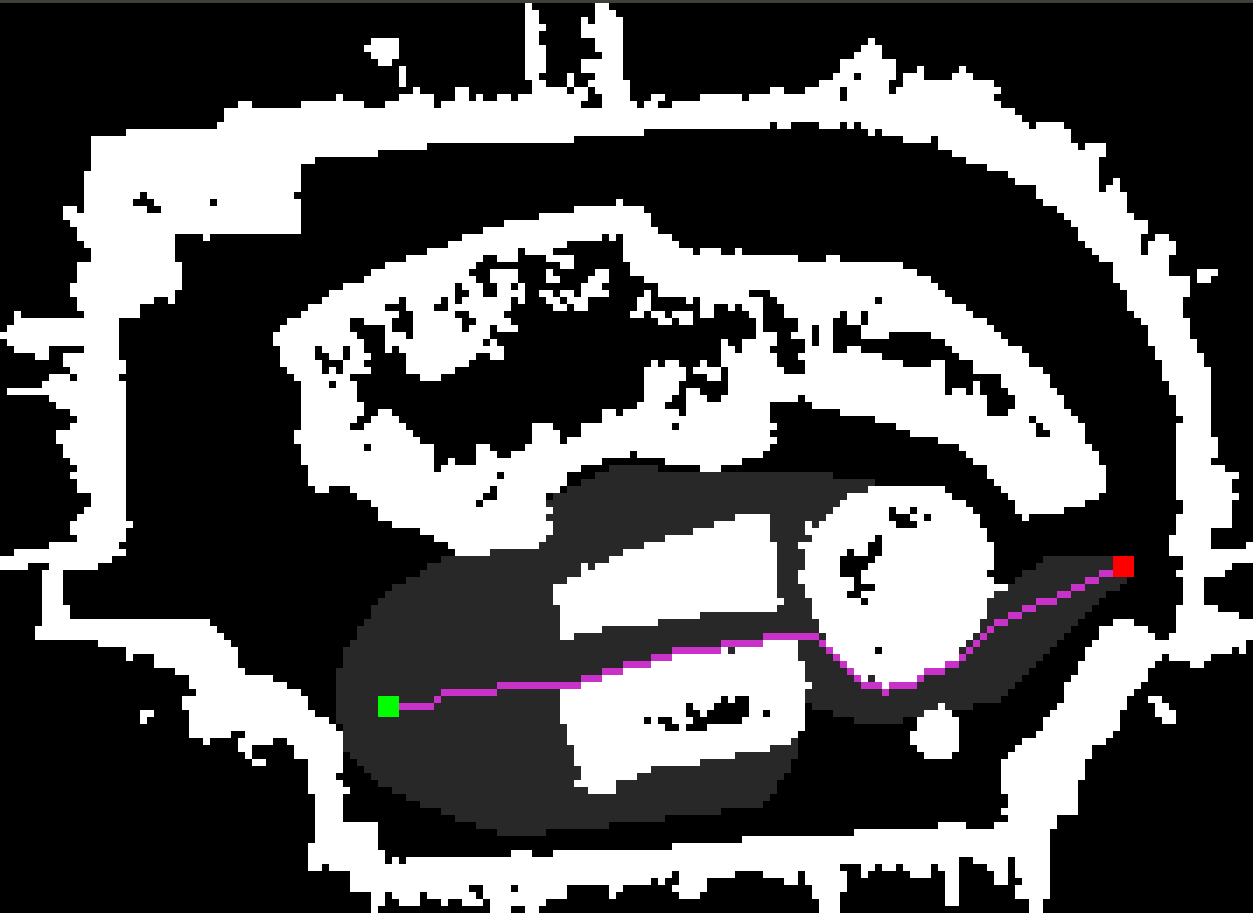}} \hspace*{0.003\textwidth}
\subfloat[]{\includegraphics[width=0.23\textwidth]{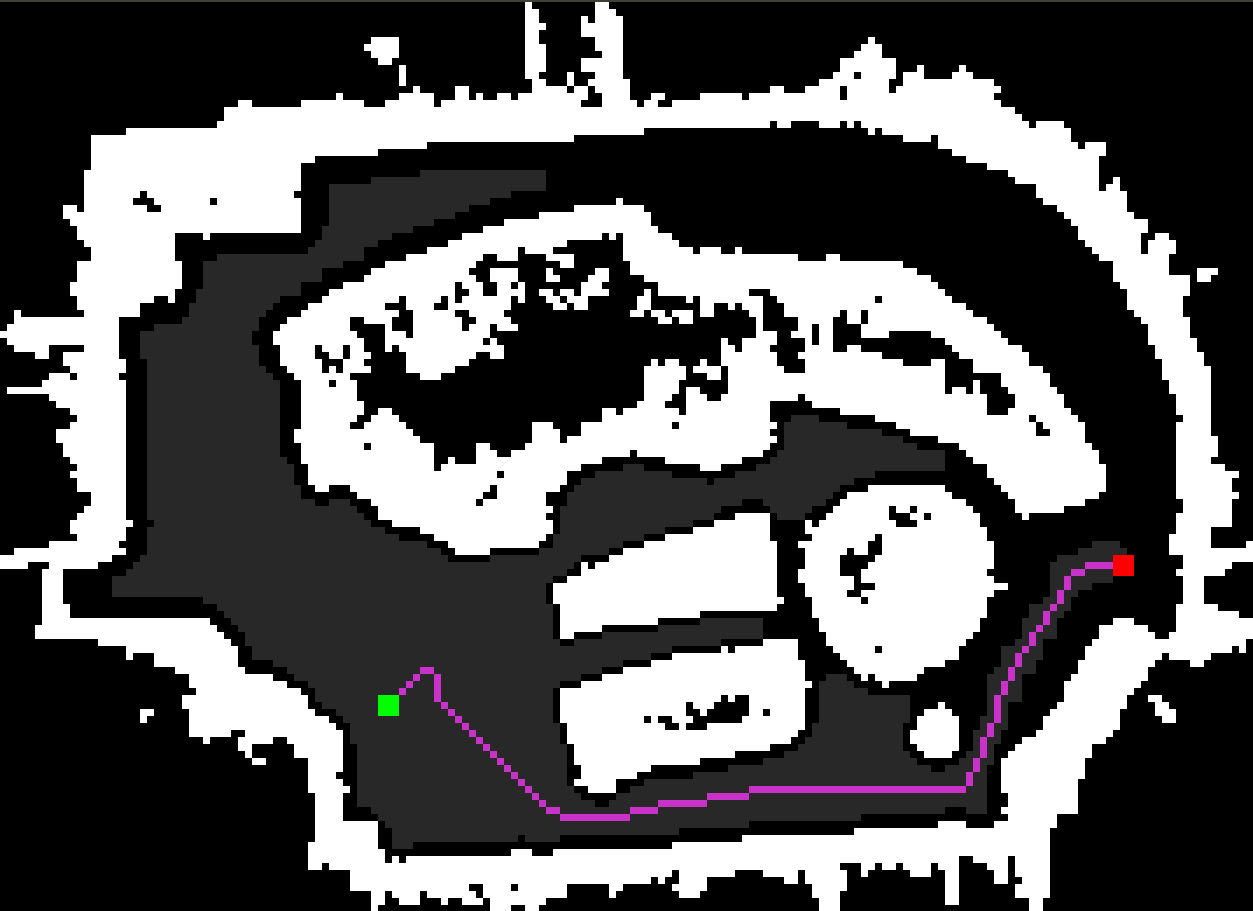}} 
    \caption{Path generated by the global path planner. The start node and goal node are correspondingly shown in green and red. (a) Original A* algorithm. (b) Modified A* algorithm with the distance cost.}
\label{fig:Astar_path}
\end{figure}

We compared the generated paths from the original A* algorithm and the modified method. These results are shown in Fig.~\ref{fig:Astar_path}. The original A* algorithm generates a path close to obstacles, whereas the modified method generates a smoothed path with keeping distances from the adjacent obstacles.

\subsection{Training neural network navigation policy}
Before actually training the policy network, a proper value of the parameter $H$ (the number of history time steps for the model input) should carefully be determined. If the given trajectory is too long, it will cause overfitting. In contrast, if the history is too short, it becomes difficult for the model to find the optimal policy. We collected a human demonstration dataset in a simulated environment that included static and dynamic obstacles. We used Gazebo \cite{koenig2004design} simulation, and we implemented the hardware and driving characteristics of DPoom in the simulation. The RGB-D camera specifications are described on the Gazebo plugin for a realistic simulation. The demonstration data were collected in environments containing $[0, 1, \dots 5]$ moving obstacles and $[0, 3, 5]$ static obstacles.\footnote{The dataset is available in: \url{ https://github.com/SeunghyunLim/Dpoom_gazebo}} Moving obstacles used the DPoom 3D model as well and were controlled by ORCA \cite{van2011reciprocal}.

\begin{figure}[t]
\centering
\vspace*{0.1in}
\includegraphics[width=0.9\linewidth]{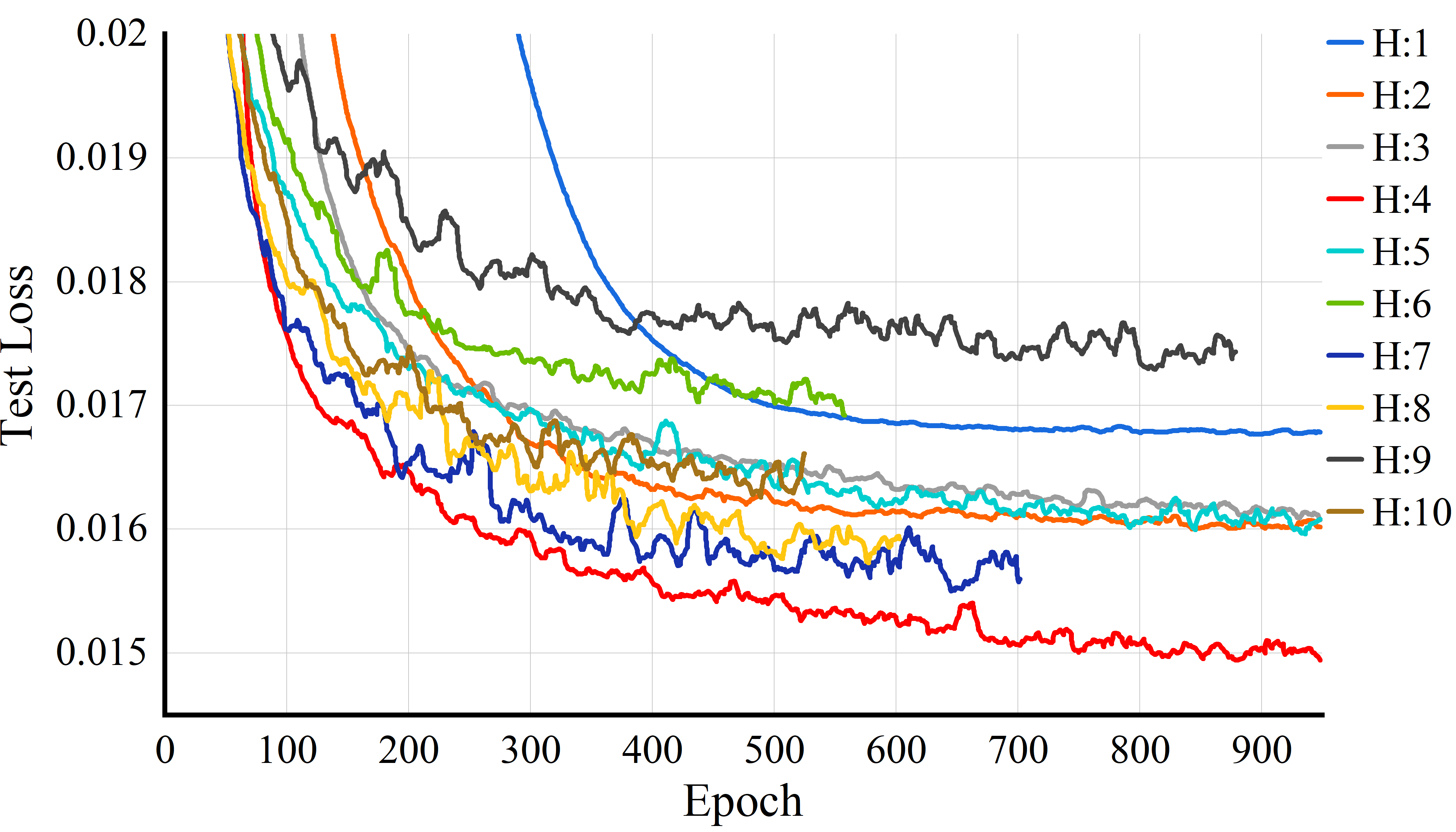}
\caption{Training results of different $H$ values.}
\label{fig:testloss}
\vspace*{-0.1in}
\end{figure}

We trained the neural network policy using the dataset with different $H$ values. We used the mean squared error (MSE) loss and Adam optimization fotraining. The results are shown in Fig.~\ref{fig:testloss}, demonstrating the fastest convergence speed and lowest test loss when $H=4$ (in red). 

For further training, we used the DAgger \cite{ross2011reduction} method, which is a basic on-policy approach of imitation learning \cite{osa2018algorithmic}. During the training procedure, we leveraged the previously aggregated dataset as the initial dataset as a warm start. Moreover, we used the pre-trained policy network as the initial policy for bootstrapping rather than using a randomly initialized policy. This strategy can also be found in recent studies of imitation learning and its effectiveness has been proven \cite{sasaki2018sample, jena2020augmenting}. The best policy model during training is saved and used in the experiments.



\subsection{Fast raw depth image ground segmentation and robot navigation}
In this section, we denote 'MORP-RB' as our rule-based navigation policy and denote 'MORP-IL' as our neural network policy $\hat{\pi}$ trained in the imitation learning manner, coupled with ground segmentation. 

\subsubsection{Training existing policies}
We implemented several existing state-of-the-art navigation methods in Gazebo for a comparison: ORCA \cite{van2011reciprocal}, CADRL \cite{chen2017decentralized}, and SARL with Local Map \cite{chen2019crowd}. The obstacles were detected with the RGB-D camera and fed into the policies. The motion commands of the policy were published to the integrated system via ROS to actuate the motors. The DRL policies designed for navigation in dynamic environments, in this case CADRL and SARL, were trained using parameters suggested by the authors \cite{chen2017decentralized, chen2019crowd}. The limited FOV and depth range of the RGB-D camera were applied to the observable area of the agents. In cases where no object was detected by the DRL agents, we fed a dummy pedestrian with a zero velocity and radius into the network, which did not affect the navigation \cite{liu2020robot}. ORCA served as the policies of the moving obstacles.

\begin{figure}[t] 
\centering
\subfloat[Gazebo view]{\includegraphics[width=0.23\textwidth]{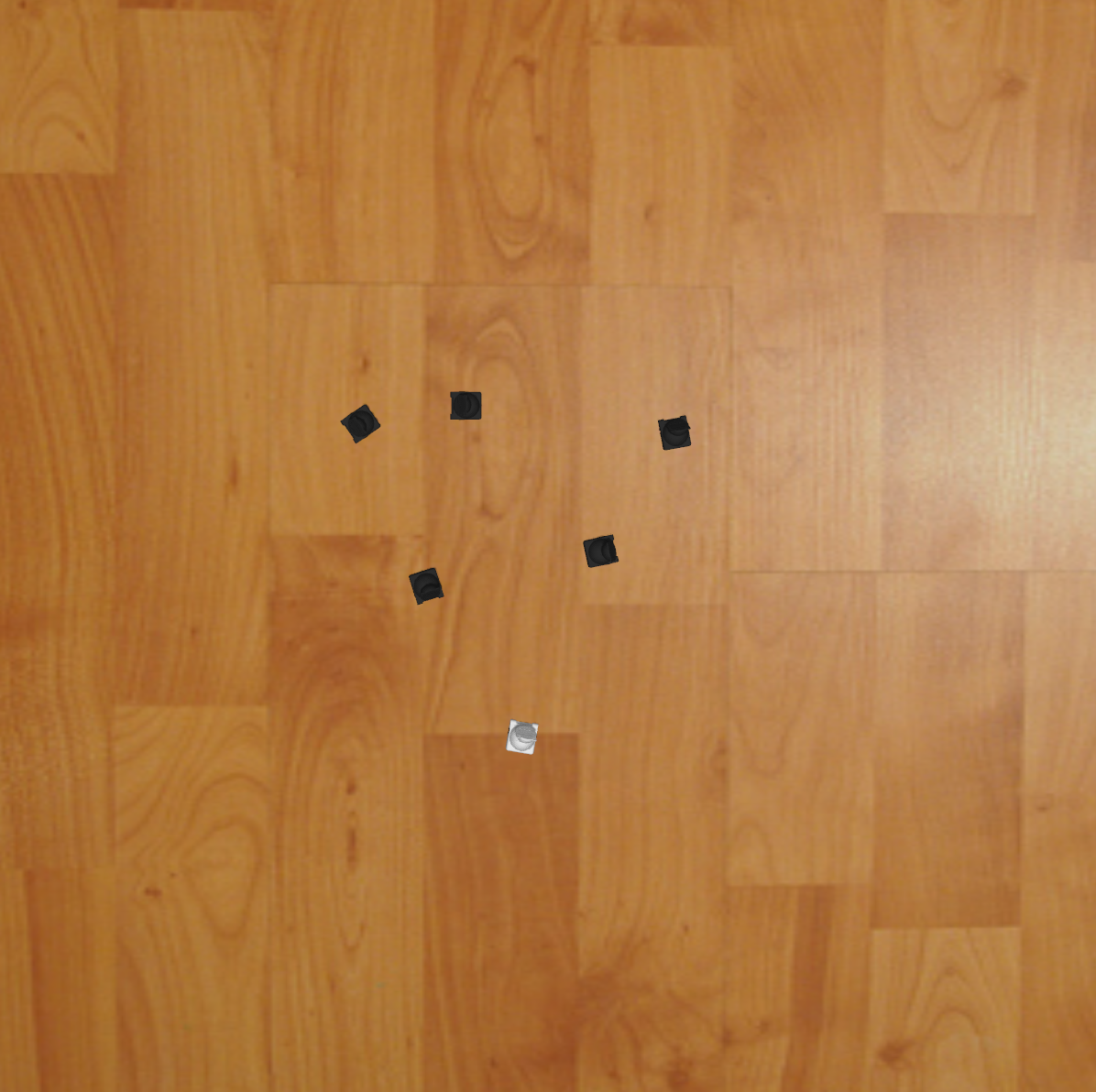}} \hspace*{0.01\textwidth}
\subfloat[ORCA]{\includegraphics[width=0.23\textwidth]{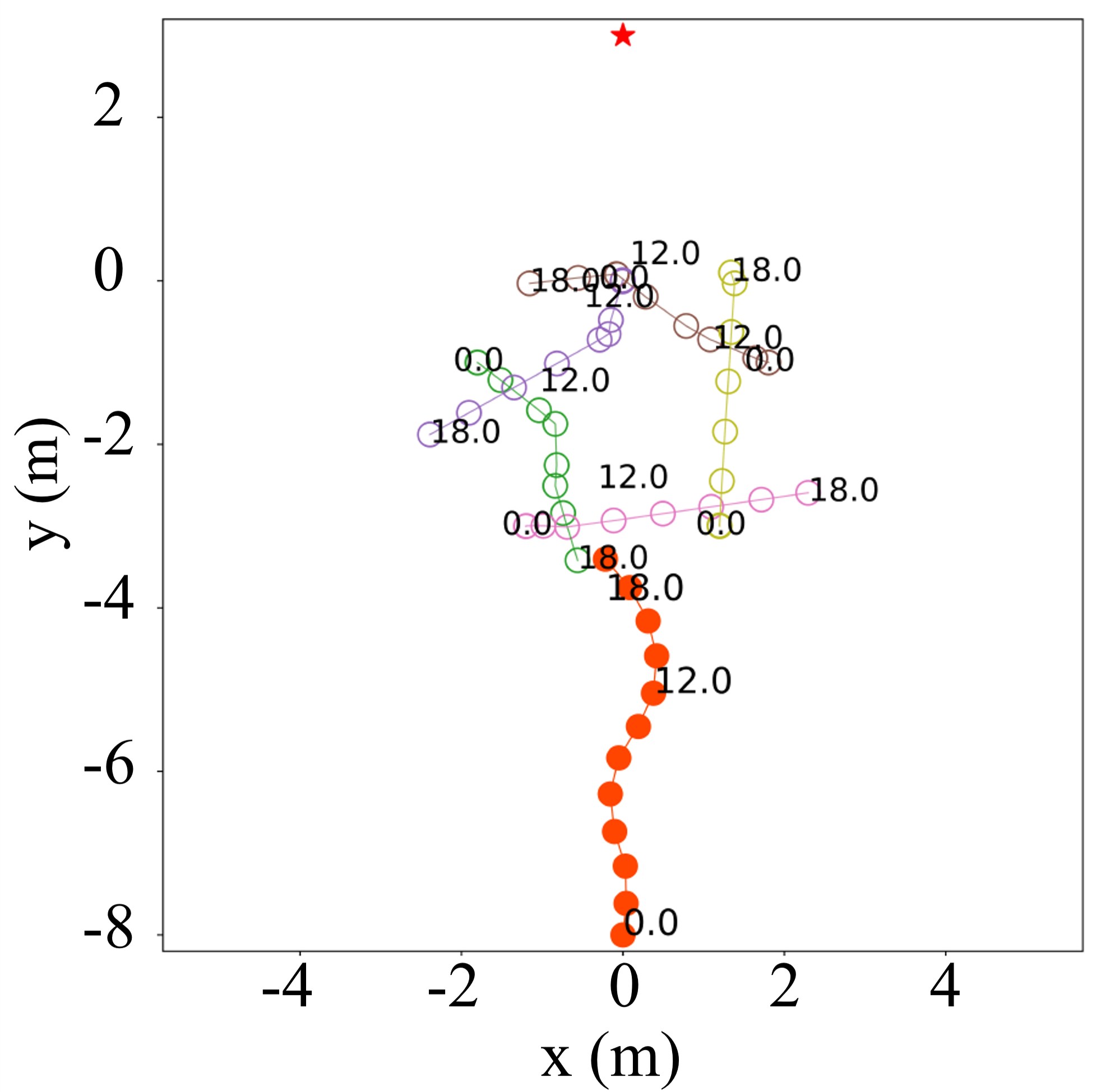}} \hfill
\subfloat[CADRL]{\includegraphics[width=0.23\textwidth]{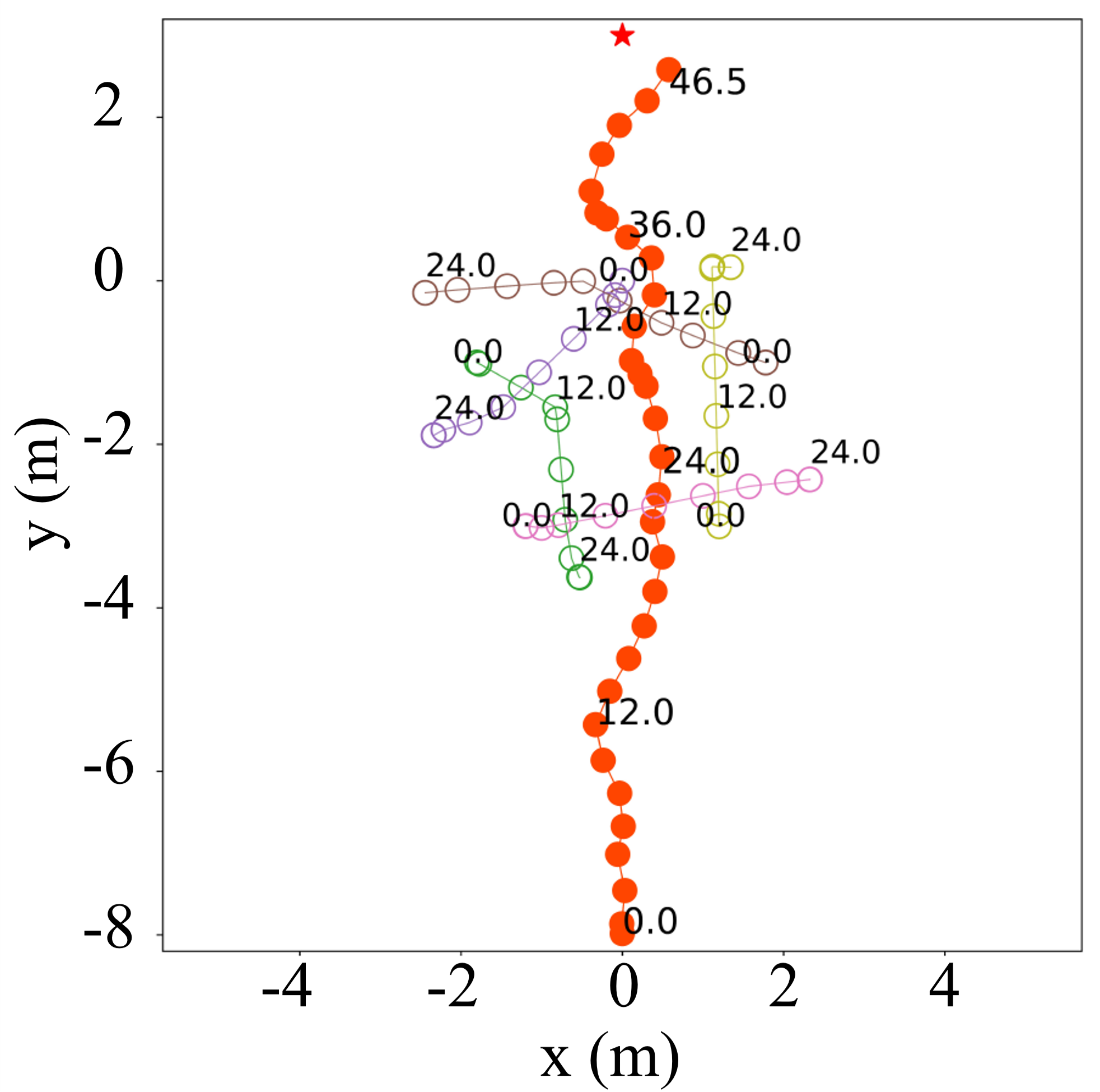}} \hspace*{0.01\textwidth}
\subfloat[SARL]{\includegraphics[width=0.23\textwidth]{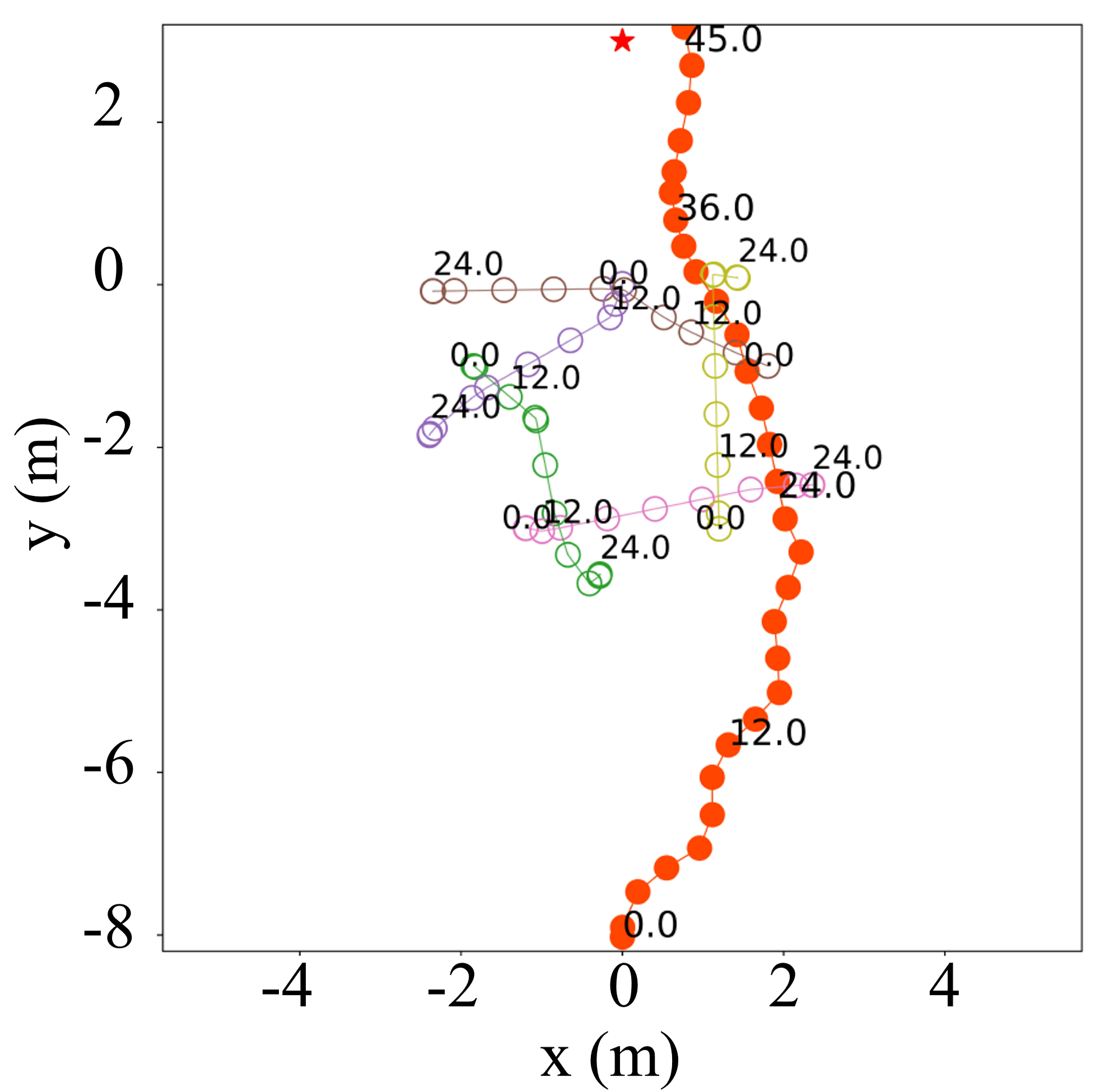}} \hfill
\subfloat[MORP-RB]{\includegraphics[width=0.23\textwidth]{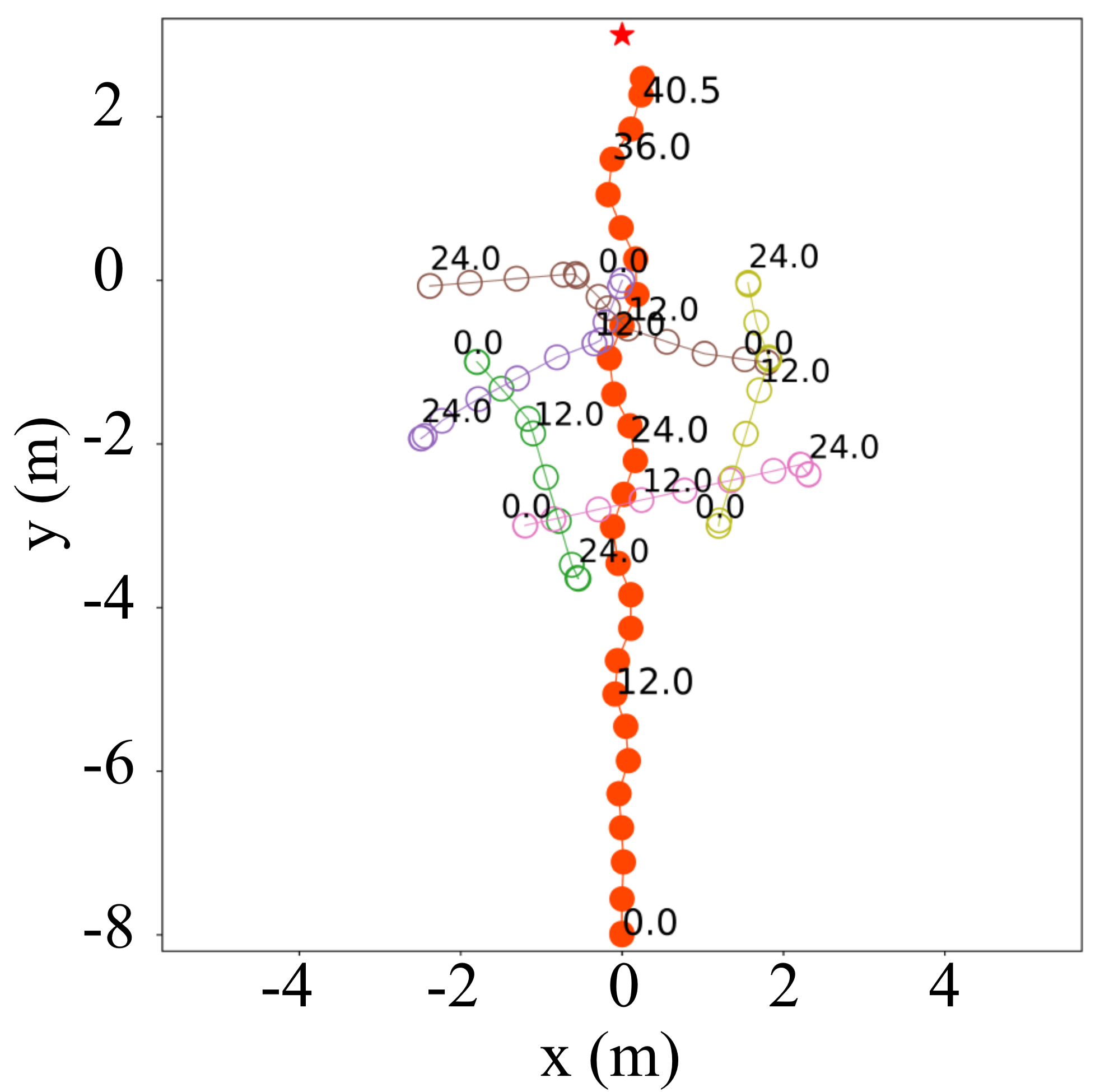}} \hspace*{0.01\textwidth}
\subfloat[MORP-IL]{\includegraphics[width=0.23\textwidth]{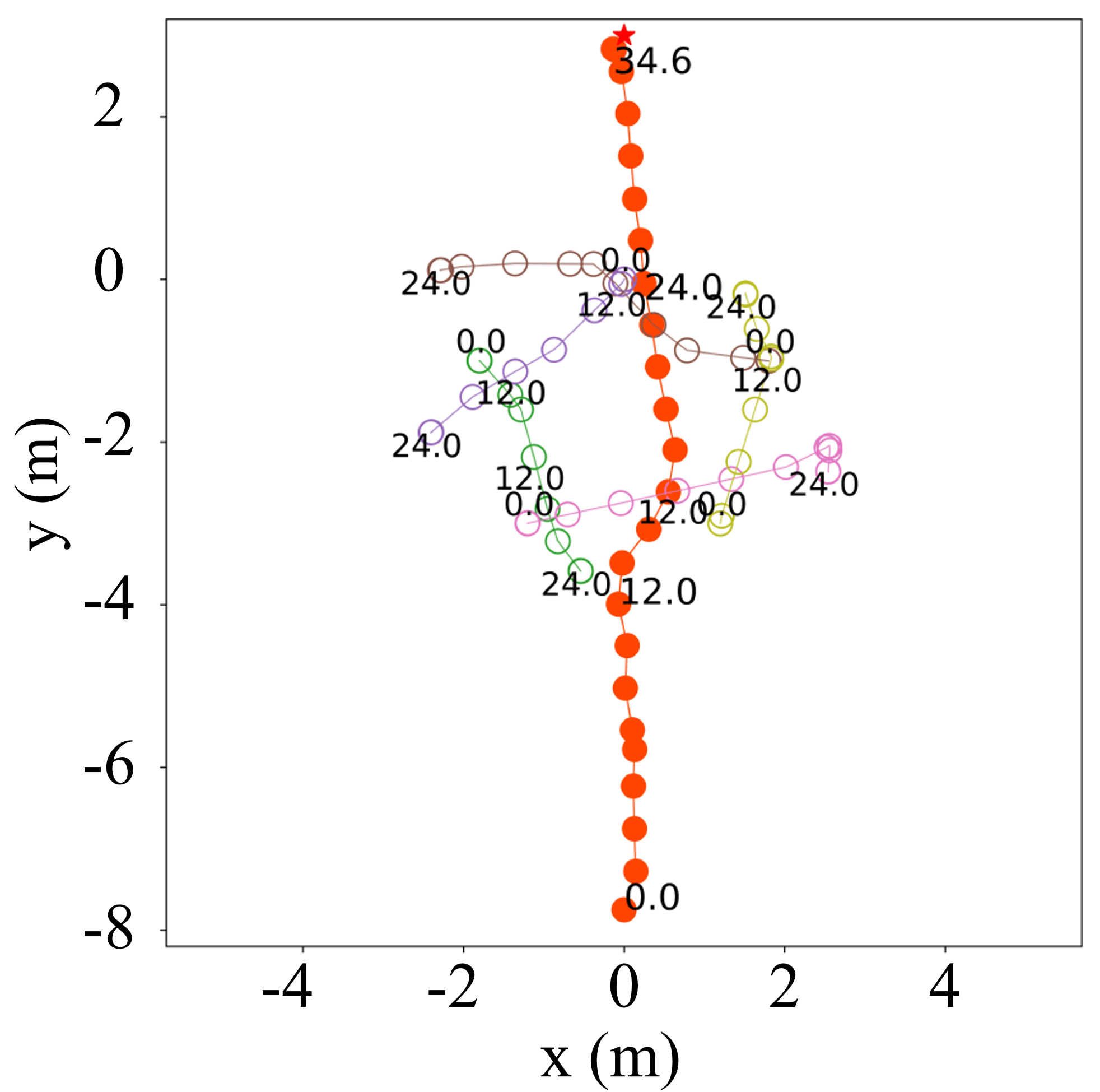}} \hfill
    \caption{Example trajectories of the compared methods without static obstacles. The robot trajectory is shown in orange.}
\label{fig:lpp_dynamic}
\end{figure}

\begin{figure}[t] 
\centering
\subfloat[Gazebo view]{\includegraphics[width=0.23\textwidth]{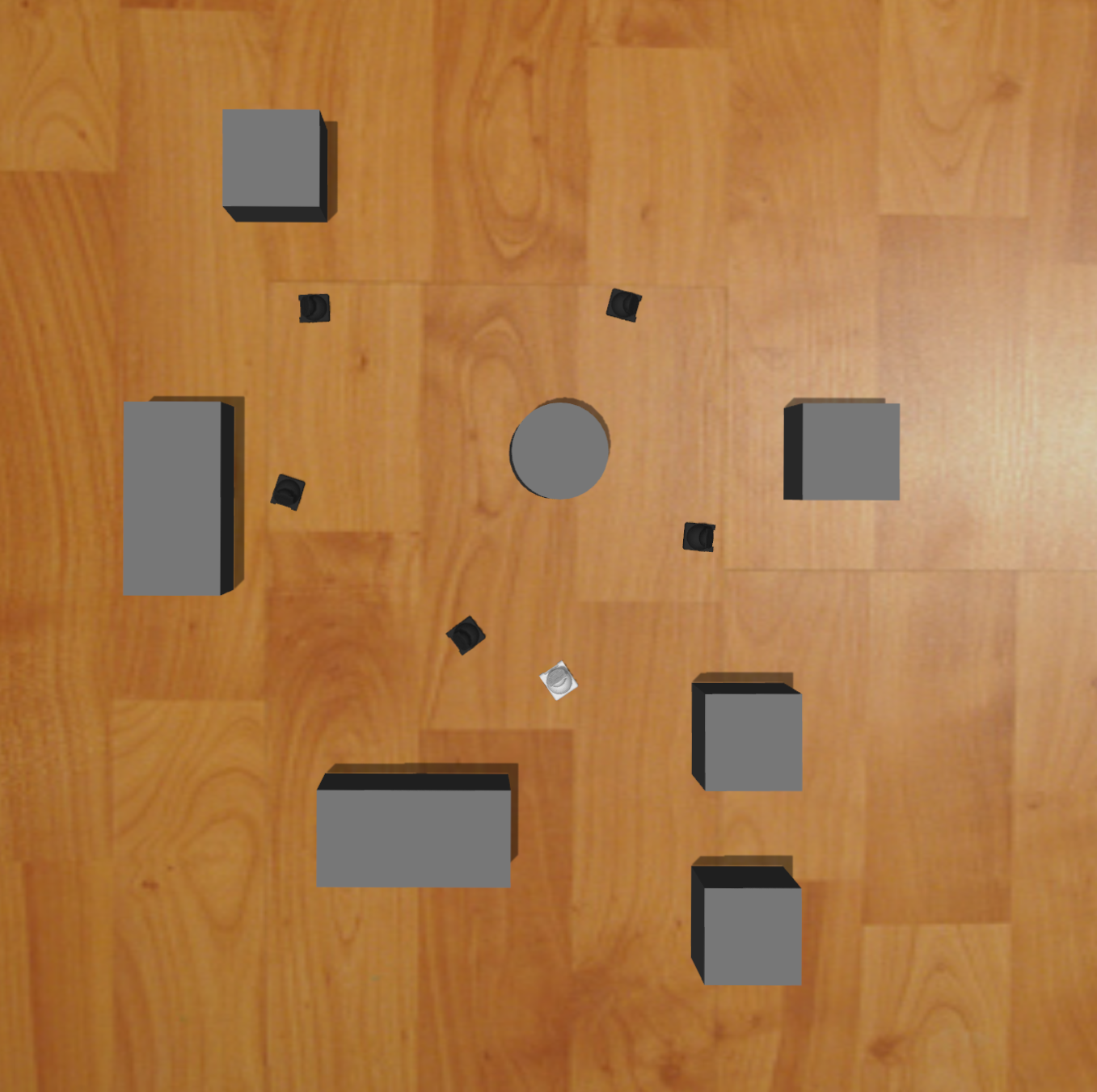}} \hspace*{0.01\textwidth}
\subfloat[ORCA]{\includegraphics[width=0.23\textwidth]{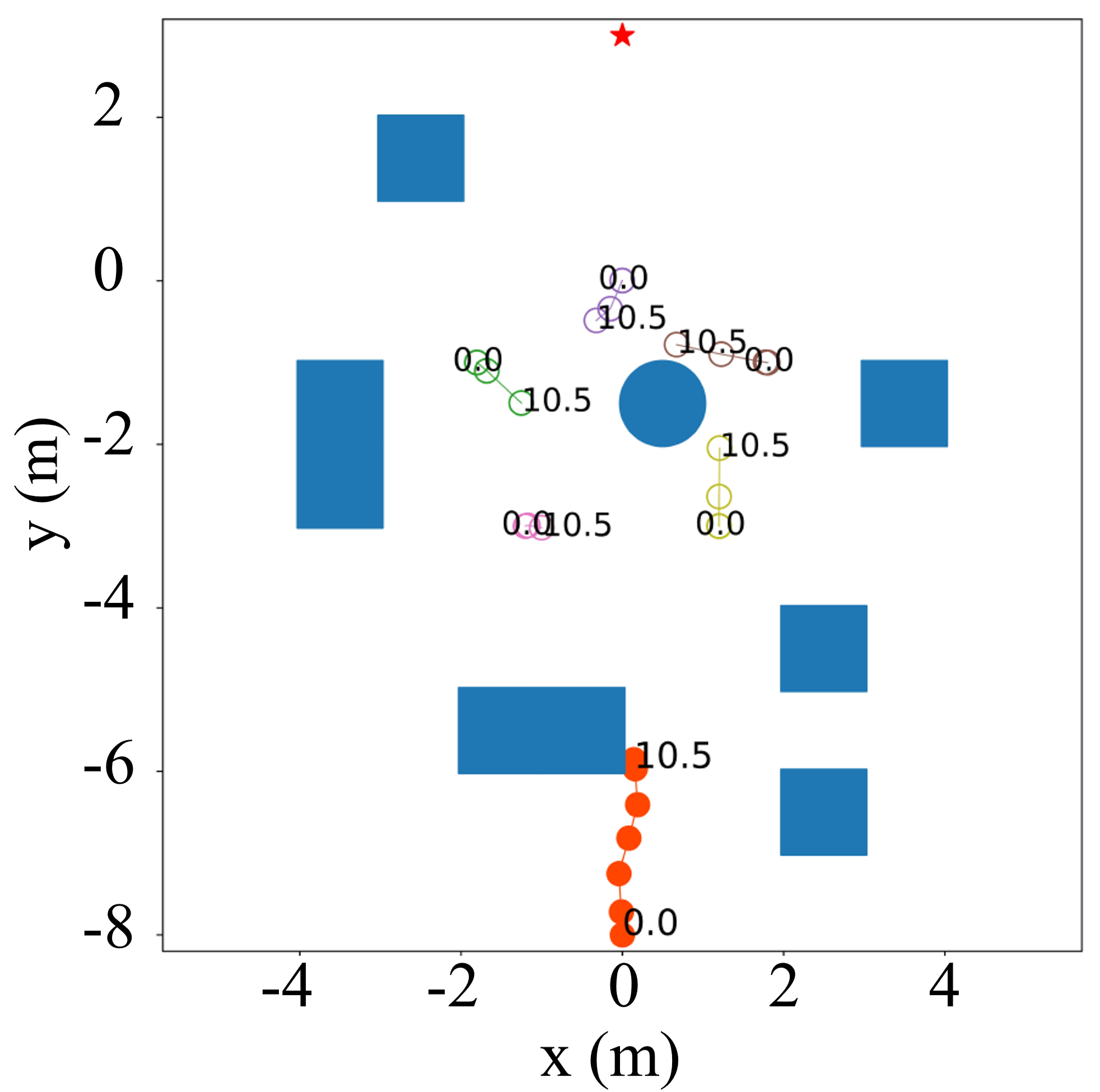}} \hfill
\subfloat[CADRL]{\includegraphics[width=0.23\textwidth]{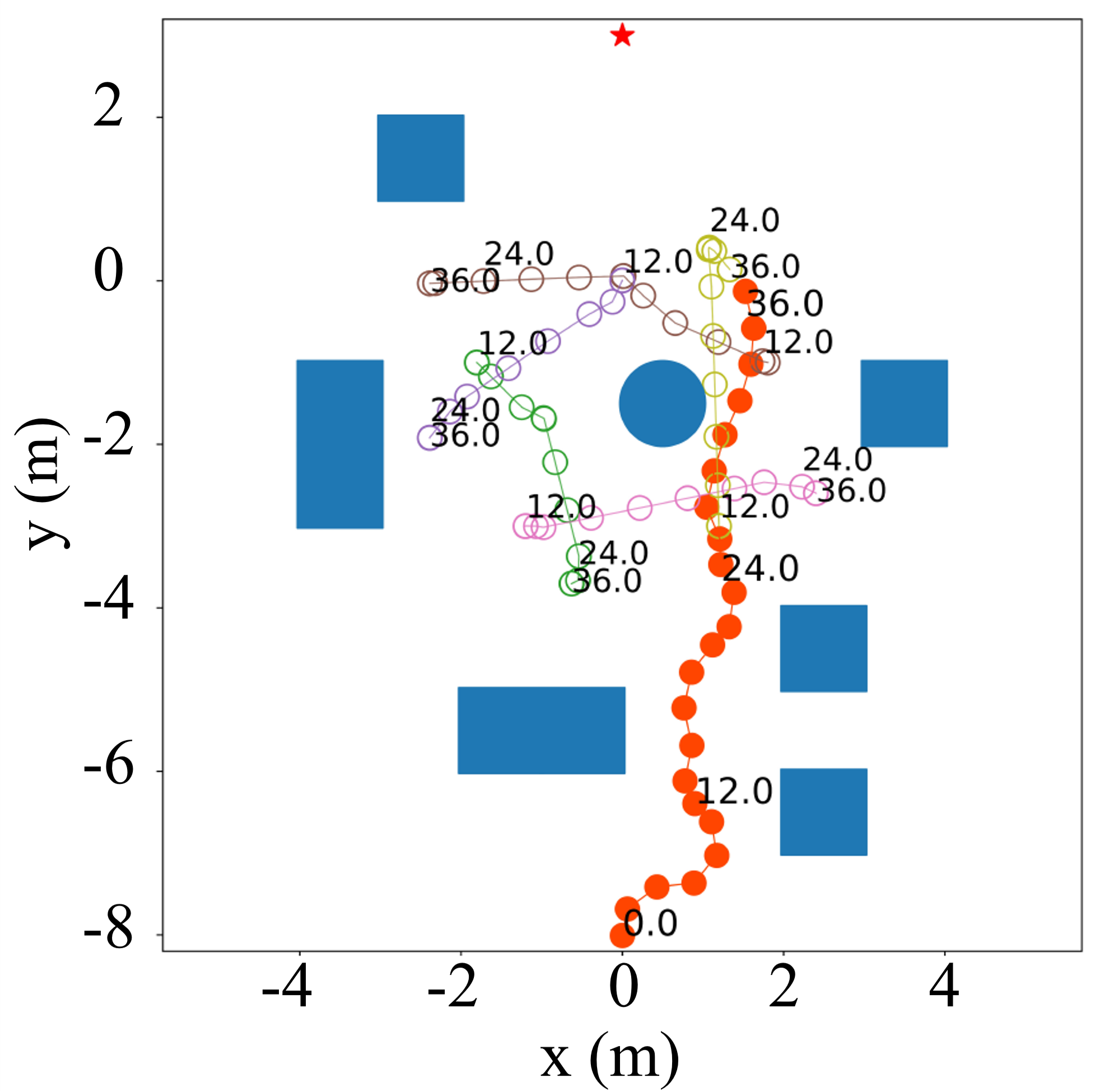}} \hspace*{0.01\textwidth}
\subfloat[SARL]{\includegraphics[width=0.23\textwidth]{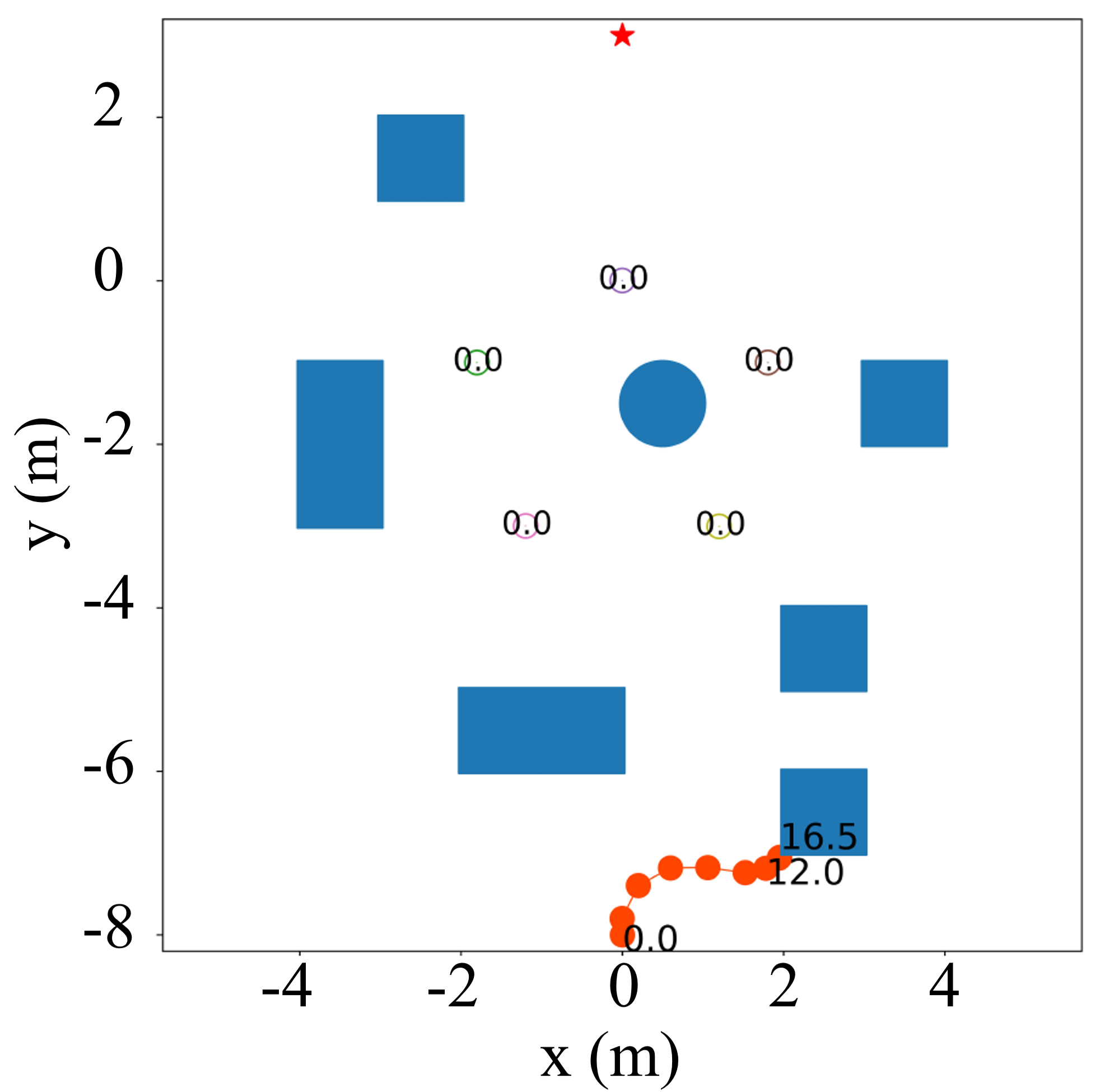}} \hfill 
\subfloat[MORP-RB]{\includegraphics[width=0.23\textwidth]{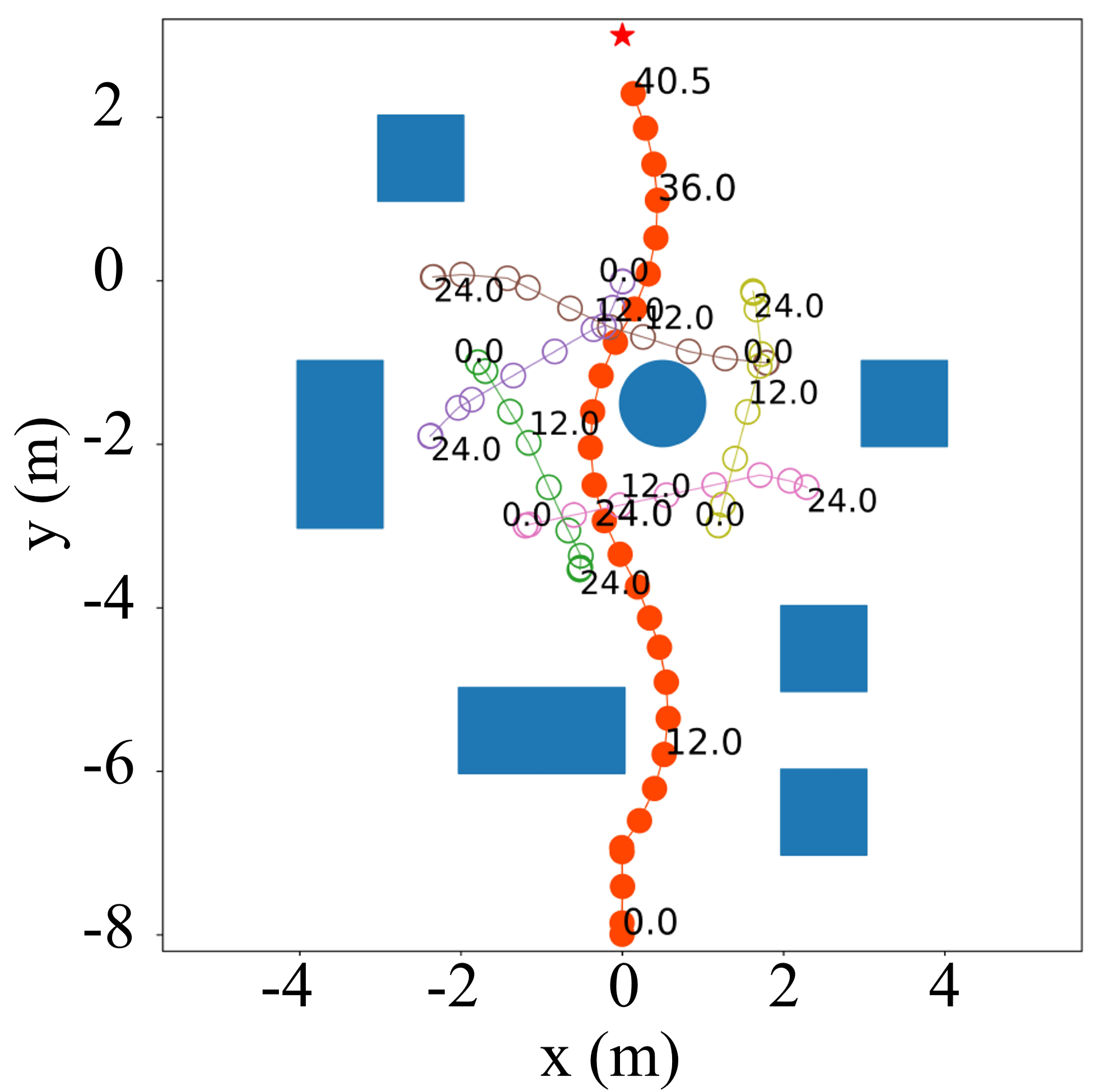}} \hspace*{0.01\textwidth}
\subfloat[MORP-IL]{\includegraphics[width=0.23\textwidth]{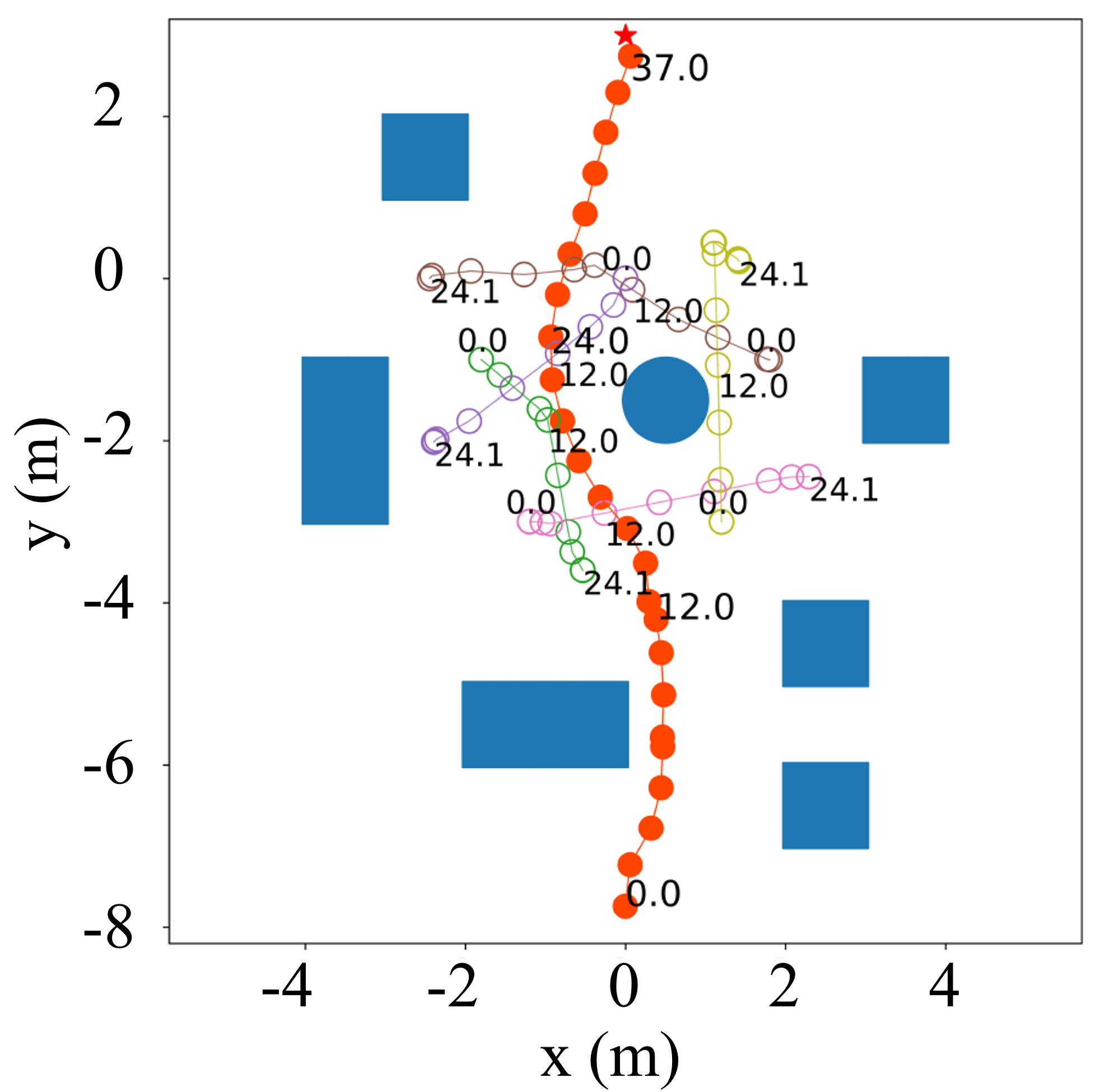}} \hfill
    \caption{Example trajectories of the compared methods with static obstacles. The static obstacles are shown in blue.}
\label{fig:lpp_static}
\end{figure}

\subsubsection{Gazebo simulation results}
The point cloud of the RGB-D camera was downsampled by voxel grid filtering and detected obstacles were fed into the ORCA and DRL policies as observed inputs. We evaluated the runtime of each navigation method by measuring the one-step time to return a motion command. We specified the runtime as three parts: the communication delay on ROS, the depth data pre-processing time (abbreviated as ``Depth."), and the decision delay of the navigation policy (abbreviated as ``Nav."). It was tested on a laptop with an Intel Core i7-8565U CPU. The results are the average of more than 100 iterations. Measurements of our methods showed that it is up to 20 times faster than the other methods compared here (see Table~\ref{table:time_took}). Unlike existing methods that require CPU and memory-intensive point cloud pre-processing, our approaches use an exteroception optimized for indoor mobile robots, greatly reducing the processing time. The compact information after ground segmentation also reduces the complexity of motion planning.

\begin{table}[h]
\vspace*{0.1in}
\renewcommand\arraystretch{1.15}
\captionsetup{justification=centering}
\caption{}
\caption*{\textsc{Runtime evaluation of different methods. All methods except for MORP-RB and MORP-IL need point cloud filtering and processing to detect obstacles. MORP-RB and MORP-IL use our fast ground segmentation.}}
\label{table:time_took}
\setlength{\tabcolsep}{3pt}
\begin{center}
{\footnotesize
\begin{tabular}{c|cccc|c}
\hline
Method & ROS delay [s] & Depth. [s] & Nav. [s] & Total [s] & Ratio\\ \hline
ORCA      & 0.00072    & 0.77477      &  0.00614 & 0.78162  & 13.84748   \\
CADRL     & 0.00072    & 0.77477      &  0.28665 & 1.06213  & 18.81710   \\
SARL      & 0.00072    & 0.77477      & 0.40651  & 1.18200  & 20.94069   \\
MORP-RB      & 0.00072    & \textbf{0.05569}      &  \textbf{0.00003}  & \textbf{0.05645} & \textbf{1.00000} \\
MORP-IL      & 0.00072    & \textbf{0.05569}      & 0.00144    & 0.05785 & 1.02486 \\
\hline
\end{tabular}
}
\end{center}
\vspace*{-0.1in}
\end{table}

We compared the following metrics for a performance evaluation: the success rate, collision rate, and average time to reach the goal. Tests were conducted separately in two different randomized environments with and without static obstacles. Goal distances from our robot were sampled from [7, 11] \textit{m}. The first environment had only pedestrians and no static obstacles. Moving obstacles used DPoom 3D models of identical sizes. Their start and goal points were empirically determined to avoid collisions with each other. There were one to five moving obstacles that were visible to all agents, but they were not able to perceive our robot (see Fig.~\ref{fig:lpp_dynamic}a). In the second environment, a unit cube and cylinder-shaped obstacles each with a 0.5 \textit{m} radius were added to the first environment, and the number of obstacles was varied from one to nine (see Fig.~\ref{fig:lpp_static}a). We modeled ten worlds for each case and ran the test three times per world. Accordingly, 60 tests were conducted for each navigation method in total.

MORP-IL shows the highest success rate while retaining a short time to reach the goal in both environments (see Tables~\ref{table:static}). The collision rate of CADRL is lower than that by our method with static obstacles because it tends to take large detours, causing it to spend twice as much time compared to the others. Apart from CADRL, our methods also show the lowest collision rate because it guarantees real-time execution. The collision rate for ORCA should be zero in an ideal 2D simulation with holonomic constraints by its design, but collisions occurred due to slow decisions by the robot and were occasionally caused by pedestrians outside of the robot's FOV. A slow refresh rate also has a disadvantage when the robot has non-holonomic constraints because the robot is unable immediately to rotate or move backwards. ORCA assumes that all pedestrians are observing the robot and avoiding it actively regardless of their FOV, which is not practical in the real world. This assumption causes the ORCA agents to take less time to reach the goal in the first environment and causes collisions with pedestrians who cannot observe the robot. Moreover, SARL has the longest one-step runtime due to its complex model architecture (see Table~\ref{table:time_took}). Akin to ORCA, slow decisions of SARL resulted in a high collision rate.

\subsubsection{Real-world experiments}
\begin{figure*}[htb] 
\centering
    \begin{subfigure}[h]{0.16\textwidth}
        \vspace*{0.1in}
        \includegraphics[width=1\linewidth]{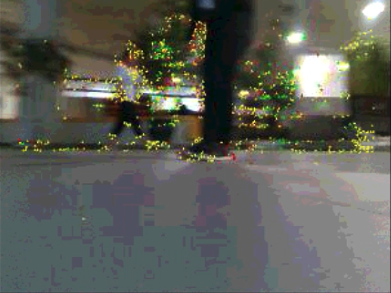}
        \label{subfig:MORP_robot_1}
    \end{subfigure}
    \begin{subfigure}[h]{0.16\textwidth}
        \vspace*{0.1in}
        \includegraphics[width=1\linewidth]{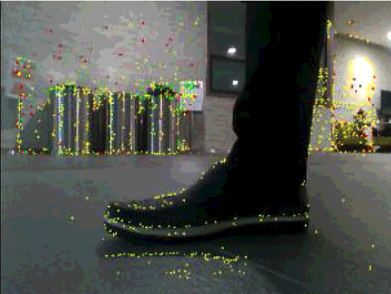}
        \label{subfig:MORP_robot_3}
    \end{subfigure}
    \begin{subfigure}[h]{0.16\textwidth}
        \vspace*{0.1in}
        \includegraphics[width=1\linewidth]{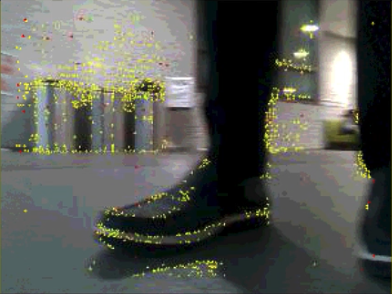}
        \label{subfig:MORP_robot_4}
    \end{subfigure}
    \begin{subfigure}[h]{0.16\textwidth}
        \vspace*{0.1in}
        \includegraphics[width=1\linewidth]{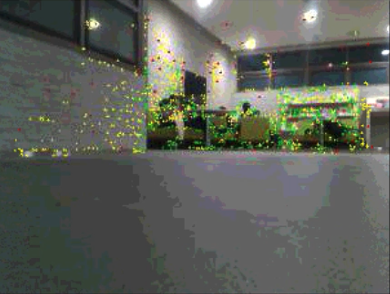}
        \label{subfig:MORP_robot_5}
    \end{subfigure}
    \begin{subfigure}[h]{0.16\textwidth}
        \vspace*{0.1in}
        \includegraphics[width=1\linewidth]{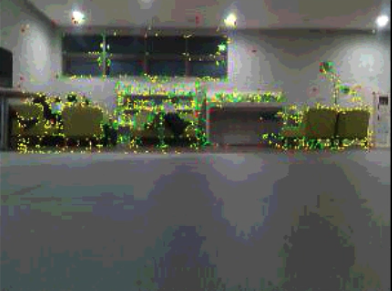}
        \label{subfig:MORP_robot_6}
    \end{subfigure}
    \begin{subfigure}[h]{0.16\textwidth}
        \vspace*{0.1in}
        \includegraphics[width=1\linewidth]{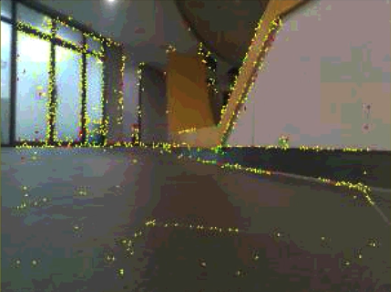}
        \label{subfig:MORP_robot_8}
    \end{subfigure}

    \begin{subfigure}[h]{0.16\textwidth}
        \includegraphics[width=1\linewidth]{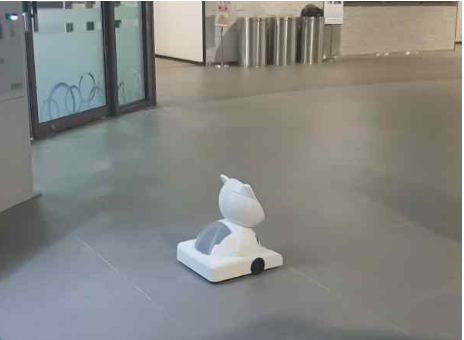}
        \caption{}
        \label{subfig:MORP_2}
        \vspace*{-0.1in}
    \end{subfigure}
    \begin{subfigure}[h]{0.16\textwidth}
        \includegraphics[width=1\linewidth]{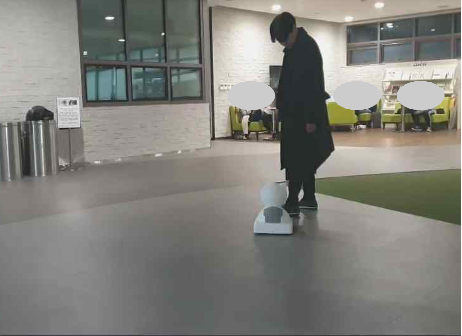}
        \caption{}
        \label{subfig:MORP_4}
        \vspace*{-0.1in}
    \end{subfigure}
    \begin{subfigure}[h]{0.16\textwidth}
        \includegraphics[width=1\linewidth]{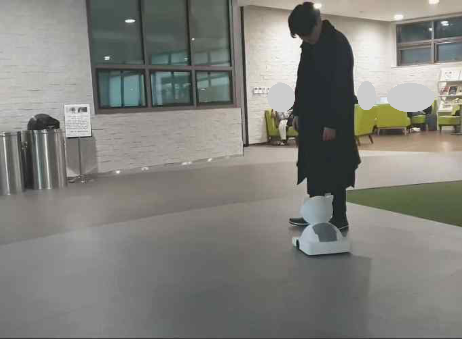}
        \caption{}
        \label{subfig:MORP_5}
        \vspace*{-0.1in}
    \end{subfigure}
    \begin{subfigure}[h]{0.16\textwidth}
        \includegraphics[width=1\linewidth]{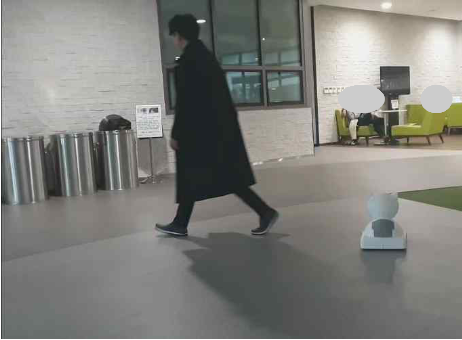}
        \caption{}
        \label{subfig:MORP_6}
        \vspace*{-0.1in}
    \end{subfigure}
    \begin{subfigure}[h]{0.16\textwidth}
        \includegraphics[width=1\linewidth]{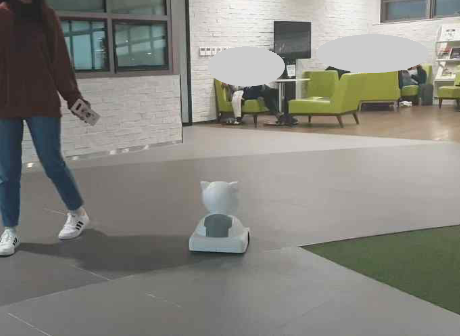}
        \caption{}
        \label{subfig:MORP_7}
        \vspace*{-0.1in}
    \end{subfigure}
    \begin{subfigure}[h]{0.16\textwidth}
        \includegraphics[width=1\linewidth]{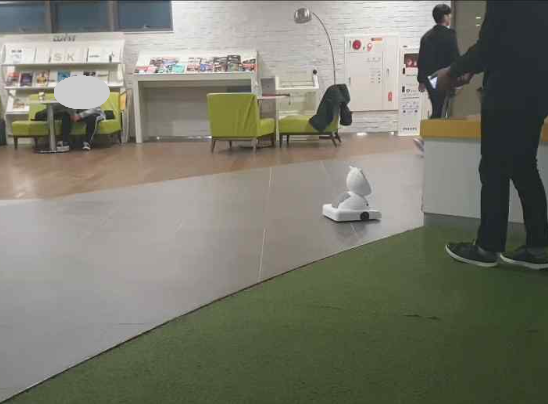}
        \caption{}
        \label{subfig:MORP_9}
        \vspace*{-0.1in}
    \end{subfigure}
    \caption{The real-world experimental result of the entire system. (a) The robot is heading to the goal location. (b)-(c) A pedestrian is found on the RGB-D camera, and the robot drives to avoid it while slowing down. (d)-(f) The robot keeps driving to the goal without global path planning, because it does not stay far from the original trajectory.}
    \label{fig:MORP}
\end{figure*}

Our navigation method was integrated into an autonomous driving system on our DPoom platform via ROS. We deployed the robot in the DGIST student dormitory lobby. For human interaction, tiny-YOLOv3 \cite{redmon2018yolov3} was used for object detection. The robot was able to estimate its pose by localization and navigate to the desired locations in the wide, crowded environment without collisions. Face emotions were displayed on the front screen depending on the situation. Our robot was able to interact in a friendly manner with people as a social robot (see Fig.~\ref{fig:MORP}).

\begin{table}[h]
\vspace*{0.1in}
\renewcommand\arraystretch{1.15}
\captionsetup{justification=centering}
\caption{}
\caption*{\textsc{Comparison of the different methods with the RGB-D camera FOV on the environment without and with static obstacles.}}
\label{table:static}
\setlength{\tabcolsep}{3pt}
\begin{center}
{\footnotesize
\begin{tabular}{c|ccc|ccc}
        & \multicolumn{3}{c|}{w/o Static Obstacles}           & \multicolumn{3}{c}{w/ Static Obstacles}             \\ \hline
Method  & Success       & Collision     & Time {[}s{]}   & Success       & Collision     & Time {[}s{]}   \\ \hline
ORCA    & 0.77          & 0.23          & \textbf{31.52} & 0.50          & 0.40          & 41.91          \\
CADRL   & 0.33          & 0.27          & 66.43          & 0.43          & \textbf{0.13} & 80.15          \\
SARL    & 0.37          & 0.10          & 45.57          & 0.30          & 0.50          & 47.52          \\
MORP-RB & \textbf{0.97} & \textbf{0.03} & 34.20          & 0.73          & 0.27          & 40.50          \\
MORP-IL & \textbf{0.97} & \textbf{0.03} & 39.49          & \textbf{0.75} & 0.25          & \textbf{38.70} \\ \hline
\end{tabular}
}
\end{center}
\vspace*{-0.1in}
\end{table}

\section{Conclusion}
\label{sec:conclusion}
In this paper, we built an open-source low-cost mobile robot platform with a single RGB-D camera. In addition, we designed a software architecture of fully autonomous navigation system for a low-cost mobile robot without LiDARs or high-end computers. For global path planning, we developed the modified A* algorithm, applying FMM to generate collision-free trajectories. For motion planning, we proposed a new RGB-D ground segmentation method that represents the traversability of the front area in the form of compact information, which is well-suited for mobile robots. This enables depth data processing in real time on a low-end SBC. We combined this idea with both rule-based and learning-based motion planners, validating that our methods can successfully navigate in crowded environments. Unlike current state-of-the-art DRL navigation approaches were slowing down while executed simultaneously with other autonomous driving functions, our approaches operated at 18Hz in real time. We also demonstrated that our methods had lower collision rates and higher success rates in a 3D simulation compared to the other methods. Finally, we deployed our autonomous driving system on our platform in a real-world residential lobby, proving the applicability of the proposed system. We tackled practical issues associated with current mobile robots and contributed to the universal use of this technology through the presentation of a price-efficient mobile robot.

\addtolength{\textheight}{0 cm}   




\bibliographystyle{IEEEtran}
\bibliography{mybib.bib}

\begin{thebibliography}{10}
\providecommand{\url}[1]{#1}
\csname url@samestyle\endcsname
\providecommand{\newblock}{\relax}
\providecommand{\bibinfo}[2]{#2}
\providecommand{\BIBentrySTDinterwordspacing}{\spaceskip=0pt\relax}
\providecommand{\BIBentryALTinterwordstretchfactor}{4}
\providecommand{\BIBentryALTinterwordspacing}{\spaceskip=\fontdimen2\font plus
\BIBentryALTinterwordstretchfactor\fontdimen3\font minus
  \fontdimen4\font\relax}
\providecommand{\BIBforeignlanguage}[2]{{%
\expandafter\ifx\csname l@#1\endcsname\relax
\typeout{** WARNING: IEEEtran.bst: No hyphenation pattern has been}%
\typeout{** loaded for the language `#1'. Using the pattern for}%
\typeout{** the default language instead.}%
\else
\language=\csname l@#1\endcsname
\fi
#2}}
\providecommand{\BIBdecl}{\relax}
\BIBdecl

\bibitem{liu2020robot}
L.~Liu, D.~Dugas, G.~Cesari, R.~Siegwart, and R.~Dub{\'e}, ``Robot navigation
  in crowded environments using deep reinforcement learning,'' in
  \emph{IEEE/RSJ International Conference on Intelligent Robots and Systems
  (IROS)}.\hskip 1em plus 0.5em minus 0.4em\relax IEEE, 2020, pp. 5671--5677.

\bibitem{mei2004energy}
Y.~Mei, Y.-H. Lu, Y.~C. Hu, and C.~G. Lee, ``Energy-efficient motion planning
  for mobile robots,'' in \emph{IEEE International Conference on Robotics and
  Automation (ICRA)}.\hskip 1em plus 0.5em minus 0.4em\relax IEEE, 2004, pp.
  4344--4349.

\bibitem{durrant2006simultaneous}
H.~Durrant-Whyte and T.~Bailey, ``Simultaneous localization and mapping: part
  i,'' \emph{IEEE Robotics \& Automation Magazine}, vol.~13, no.~2, pp.
  99--110, 2006.

\bibitem{chen2017decentralized}
Y.~F. Chen, M.~Liu, M.~Everett, and J.~P. How, ``Decentralized
  non-communicating multiagent collision avoidance with deep reinforcement
  learning,'' in \emph{IEEE International Conference on Robotics and Automation
  (ICRA)}.\hskip 1em plus 0.5em minus 0.4em\relax IEEE, 2017, pp. 285--292.

\bibitem{everett2018motion}
M.~Everett, Y.~F. Chen, and J.~P. How, ``Motion planning among dynamic,
  decision-making agents with deep reinforcement learning,'' in \emph{IEEE/RSJ
  International Conference on Intelligent Robots and Systems (IROS)}.\hskip 1em
  plus 0.5em minus 0.4em\relax IEEE, 2018, pp. 3052--3059.

\bibitem{chen2019crowd}
C.~Chen, Y.~Liu, S.~Kreiss, and A.~Alahi, ``Crowd-robot interaction:
  Crowd-aware robot navigation with attention-based deep reinforcement
  learning,'' in \emph{IEEE International Conference on Robotics and Automation
  (ICRA)}.\hskip 1em plus 0.5em minus 0.4em\relax IEEE, 2019, pp. 6015--6022.

\bibitem{Hart1968}
P.~E. Hart, N.~J. Nilsson, and B.~Raphael, ``Formal basis for the heuristic
  determination of mininmum cost paths,'' \emph{Systems Science and
  Cybernetics}, vol.~4, no.~2, pp. 100--107, 1968.

\bibitem{haddadin2009requirements}
S.~Haddadin, A.~Albu-Sch{\"a}ffer, and G.~Hirzinger, ``Requirements for safe
  robots: Measurements, analysis and new insights,'' \emph{The International
  Journal of Robotics Research}, vol.~28, no. 11-12, pp. 1507--1527, 2009.

\bibitem{van2008reciprocal}
J.~Van~den Berg, M.~Lin, and D.~Manocha, ``Reciprocal velocity obstacles for
  real-time multi-agent navigation,'' in \emph{IEEE International Conference on
  Robotics and Automation (ICRA)}.\hskip 1em plus 0.5em minus 0.4em\relax IEEE,
  2008, pp. 1928--1935.

\bibitem{van2011reciprocal}
J.~Van Den~Berg, S.~J. Guy, M.~Lin, and D.~Manocha, ``Reciprocal n-body
  collision avoidance,'' in \emph{Robotics research}.\hskip 1em plus 0.5em
  minus 0.4em\relax Springer, 2011, pp. 3--19.

\bibitem{chen2017socially}
Y.~F. Chen, M.~Everett, M.~Liu, and J.~P. How, ``Socially aware motion planning
  with deep reinforcement learning,'' in \emph{IEEE/RSJ International
  Conference on Intelligent Robots and Systems (IROS)}.\hskip 1em plus 0.5em
  minus 0.4em\relax IEEE, 2017, pp. 1343--1350.

\bibitem{yang2019robustifying}
K.~Yang, L.~M. Bergasa, E.~Romera, and K.~Wang, ``Robustifying semantic
  cognition of traversability across wearable rgb-depth cameras,''
  \emph{Applied optics}, vol.~58, no.~12, pp. 3141--3155, 2019.

\bibitem{paigwar2020gndnet}
A.~Paigwar, {\"O}.~Erkent, D.~S. Gonz{\'a}lez, and C.~Laugier, ``Gndnet: Fast
  ground plane estimation and point cloud segmentation for autonomous
  vehicles,'' in \emph{IEEE/RSJ International Conference on Intelligent Robots
  and Systems (IROS)}, 2020.

\bibitem{quigley2009ros}
M.~Quigley, K.~Conley, B.~Gerkey, J.~Faust, T.~Foote, J.~Leibs, R.~Wheeler,
  A.~Y. Ng \emph{et~al.}, ``Ros: an open-source robot operating system,'' in
  \emph{ICRA workshop on open source software}, vol.~3, no. 3.2.\hskip 1em plus
  0.5em minus 0.4em\relax Kobe, Japan, 2009, p.~5.

\bibitem{labbe2019rtab}
M.~Labb{\'e} and F.~Michaud, ``Rtab-map as an open-source lidar and visual
  simultaneous localization and mapping library for large-scale and long-term
  online operation,'' \emph{Journal of Field Robotics}, vol.~36, no.~2, pp.
  416--446, 2019.

\bibitem{Khanmirza2018}
E.~Khanmirza, M.~Haghbeigi, M.~Nazarahari, and S.~Doostie, ``{A Comparative
  Study of Deterministic and Probabilistic Mobile Robot Path Planning
  Algorithms},'' \emph{5th RSI International Conference on Robotics and
  Mechatronics (IcRoM)}, pp. 534--539, 2018.

\bibitem{dijkstra1959}
E.~W. Dijkstra, ``A note on two problems in connexion with graphs,''
  \emph{Numerische Mathematik}, vol.~1, no.~1, p. 269–271, 1959.

\bibitem{Lim2020}
S.~Lim, S.~W. Sohn, H.~Lee, D.~Choi, E.~Jang, M.~Kim, J.~Lee, and S.~Park,
  ``{Analysis and Evaluation of Path Planning Algorithms for Autonomous Driving
  of Electromagnetically Actuated Microrobot},'' \emph{International Journal of
  Control, Automation and Systems}, vol.~18, pp. 1--12, 2020.

\bibitem{Sethian2016}
J.~A. Sethian, ``{A Fast Marching Level Set Method for Monotonically Advancing
  Fronts},'' \emph{Proceedings of the National Academy of Sciences of the
  United States of America}, vol.~93, no.~4, pp. 1591--1595, 2016.

\bibitem{schilling2017geometric}
F.~Schilling, X.~Chen, J.~Folkesson, and P.~Jensfelt, ``Geometric and visual
  terrain classification for autonomous mobile navigation,'' in \emph{IEEE/RSJ
  International Conference on Intelligent Robots and Systems (IROS)}.\hskip 1em
  plus 0.5em minus 0.4em\relax IEEE, 2017, pp. 2678--2684.

\bibitem{holz2011real}
D.~Holz, S.~Holzer, R.~B. Rusu, and S.~Behnke, ``Real-time plane segmentation
  using rgb-d cameras,'' in \emph{Robot Soccer World Cup}.\hskip 1em plus 0.5em
  minus 0.4em\relax Springer, 2011, pp. 306--317.

\bibitem{himmelsbach2010fast}
M.~Himmelsbach, F.~V. Hundelshausen, and H.-J. Wuensche, ``Fast segmentation of
  3d point clouds for ground vehicles,'' in \emph{IEEE Intelligent Vehicles
  Symposium}.\hskip 1em plus 0.5em minus 0.4em\relax IEEE, 2010, pp. 560--565.

\bibitem{holz2010towards}
D.~Holz, R.~Schnabel, D.~Droeschel, J.~St{\"u}ckler, and S.~Behnke, ``Towards
  semantic scene analysis with time-of-flight cameras,'' in \emph{Robot Soccer
  World Cup}.\hskip 1em plus 0.5em minus 0.4em\relax Springer, 2010, pp.
  121--132.

\bibitem{osa2018algorithmic}
T.~Osa, J.~Pajarinen, G.~Neumann, J.~A. Bagnell, P.~Abbeel, J.~Peters
  \emph{et~al.}, ``An algorithmic perspective on imitation learning,''
  \emph{Foundations and Trends in Robotics}, vol.~7, no. 1-2, pp. 1--179, 2018.

\bibitem{nair2010rectified}
V.~Nair and G.~E. Hinton, ``Rectified linear units improve restricted boltzmann
  machines,'' in \emph{Proceedings of the International Conference on Machine
  Learning (ICML)}, 2010.

\bibitem{punjani2015deep}
A.~Punjani and P.~Abbeel, ``Deep learning helicopter dynamics models,'' in
  \emph{IEEE International Conference on Robotics and Automation (ICRA)}.\hskip
  1em plus 0.5em minus 0.4em\relax IEEE, 2015, pp. 3223--3230.

\bibitem{spielberg2019neural}
N.~A. Spielberg, M.~Brown, N.~R. Kapania, J.~C. Kegelman, and J.~C. Gerdes,
  ``Neural network vehicle models for high-performance automated driving,''
  \emph{Science Robotics}, vol.~4, no.~28, 2019.

\bibitem{kumar2021rma}
A.~Kumar, Z.~Fu, D.~Pathak, and J.~Malik, ``Rma: Rapid motor adaptation for
  legged robots,'' \emph{arXiv preprint arXiv:2107.04034}, 2021.

\bibitem{koenig2004design}
N.~Koenig and A.~Howard, ``Design and use paradigms for gazebo, an open-source
  multi-robot simulator,'' in \emph{IEEE/RSJ International Conference on
  Intelligent Robots and Systems (IROS)}.\hskip 1em plus 0.5em minus
  0.4em\relax IEEE, 2004, pp. 2149--2154.

\bibitem{ross2011reduction}
S.~Ross, G.~Gordon, and D.~Bagnell, ``A reduction of imitation learning and
  structured prediction to no-regret online learning,'' in \emph{Proceedings of
  the International Conference on Artificial Intelligence and Statistics
  (AISTAT)}, 2011, pp. 627--635.

\bibitem{sasaki2018sample}
F.~Sasaki, T.~Yohira, and A.~Kawaguchi, ``Sample efficient imitation learning
  for continuous control,'' in \emph{Proceedings of the International
  Conference on Learning Representations (ICLR)}, 2018.

\bibitem{jena2020augmenting}
R.~Jena, C.~Liu, and K.~Sycara, ``Augmenting gail with bc for sample efficient
  imitation learning,'' \emph{arXiv preprint arXiv:2001.07798}, 2020.

\bibitem{redmon2018yolov3}
J.~Redmon and A.~Farhadi, ``Yolov3: An incremental improvement,'' \emph{arXiv
  preprint arXiv:1804.02767}, 2018.

\end{thebibliography}

\end{document}